\documentclass[lettersize,journal]{IEEEtran}
\usepackage{amsmath,amsfonts}
\usepackage{algorithmic}
\usepackage{array}
\usepackage{textcomp}
\usepackage{stfloats}
\usepackage{url}
\usepackage{verbatim}
\usepackage{graphicx}
\usepackage{amssymb}
\usepackage{algorithm}
\usepackage{booktabs}
\usepackage[table]{xcolor}
\usepackage{multirow}
\usepackage{bbding}
\usepackage{pifont}
\usepackage{wasysym}
\usepackage{utfsym}
\usepackage{fontawesome}
\usepackage{array}
\usepackage[colorlinks,
            linkcolor=blue,
            anchorcolor=blue,
            citecolor=blue]{hyperref}
\usepackage[nocompress]{cite}
\usepackage{textcomp}
\usepackage{stfloats}
\usepackage{url}
\usepackage{verbatim}
\usepackage{graphicx}
\usepackage{cite}
\usepackage[caption=false,font=normalsize,labelfont=sf,textfont=sf]{subfig}
\def\BibTeX{{\rm B\kern-.05em{\sc i\kern-.025em b}\kern-.08em
    T\kern-.1667em\lower.7ex\hbox{E}\kern-.125emX}}
    
\usepackage{balance}

\begin{document}
\title{WaveSeg: Enhancing Segmentation Precision via High-Frequency Prior and Mamba-Driven Spectrum Decomposition}

\author{Guoan Xu, 
        Yang Xiao, 
        Wenjing Jia,~\IEEEmembership{Member,~IEEE}, 
        Guangwei Gao,~\IEEEmembership{Senior Member,~IEEE},
        Guo-Jun Qi,~\IEEEmembership{Fellow,~IEEE},
        and Chia-Wen Lin,~\IEEEmembership{Fellow,~IEEE}

\thanks{Guoan Xu, Yang Xiao, and Wenjing Jia are with the Faculty of Engineering and Information Technology, University of Technology Sydney, Sydney, NSW 2007, Australia (e-mail: xga\_njupt@163.com, yang.xiao-2@student.uts.edu.au, and Wenjing.Jia@uts.edu.au).}


\thanks{Guangwei Gao is with the PCA Laboratory, the Key Laboratory of Intelligent Perception and Systems for High-Dimensional Information of the Ministry of Education, School of Computer Science and Engineering, Nanjing University of Science and Technology, Nanjing 210094, China (e-mail: csggao@gmail.com).}

\thanks{Guo-Jun Qi is with the Research Center for Industries of the Future and the School of Engineering, Westlake University, Hangzhou 310024, China, and also with OPPO Research, Seattle, WA 98101 USA (e-mail: guojunq@gmail.com).}

\thanks{Chia-Wen Lin is with the Department of Electrical Engineering and the Institute of Communications Engineering, National Tsing Hua University, Hsinchu 300044, Taiwan (e-mail: cwlin@ee.nthu.edu.tw).}
}

\markboth{Journal of \LaTeX\ Class Files,~Vol.~14, No.~8, August~2021}%
{Shell \MakeLowercase{\textit{et al.}}: A Sample Article Using IEEEtran.cls for IEEE Journals}

\maketitle

\begin{abstract}
While recent semantic segmentation networks heavily rely on powerful pretrained encoders, most employ simplistic decoders, leading to suboptimal trade-offs between semantic context and fine-grained detail preservation. To address this, we propose a novel decoder architecture, WaveSeg, which jointly optimizes feature refinement in spatial and wavelet domains. Specifically, high-frequency components are first learned from input images as explicit priors to reinforce boundary details at early stages. A multi-scale fusion mechanism, Dual Domain Operation (DDO), is then applied, and the novel Spectrum Decomposition Attention (SDA) block is proposed, which is developed to leverage Mamba’s linear-complexity long-range modeling to enhance high-frequency structural details. Meanwhile, reparameterized convolutions are applied to preserve low-frequency semantic integrity in the wavelet domain. 
Finally, a residual-guided fusion integrates multi-scale features with boundary-aware representations at native resolution, producing semantically and structurally rich feature maps.
Extensive experiments on standard benchmarks demonstrate that WaveSeg, leveraging wavelet-domain frequency prior with Mamba-based attention, consistently outperforms state-of-the-art approaches both quantitatively and qualitatively, achieving efficient and precise segmentation.

\end{abstract}

\begin{IEEEkeywords}
Semantic segmentation, semantic context, fine-grained details, mamba, wavelet domain.
\end{IEEEkeywords}

\section{Introduction}
\IEEEPARstart{S}{emantic} segmentation is a core task in computer vision, aimed at assigning a specific label to each pixel in an image. This technique is widely applied in various fields, such as autonomous driving~\cite{siam2017deep}, medical imaging~\cite{butoi2023universeg}, and robotics~\cite{colleoni2021robotic}. Despite these advances, achieving precise pixel-level predictions remains a challenge. Modern semantic segmentation models~\cite{xie2021segformer,guo2022segnext,ni2024context,yu2025rethinking} are typically composed of encoder-decoder architectures. However, most existing methods predominantly focused on designing powerful backbones, while the decoder, though playing a critical role in aggregating and refining multi-scale features, is often overlooked. Although the introduction of hierarchical backbones~\cite{wang2021pyramid,xie2021segformer,yu2025mambaout}, which produce multi-scale features, 
has significantly alleviated the challenge of modeling both global and local pixel relationships, many segmentation frameworks still adopt simplistic decoders based on multilayer perceptrons (MLPs) or convolutional blocks and often fall short in fully leveraging images' rich hierarchical features for precise prediction.  




\begin{figure}[t]
   \centering
   \begin{minipage}[b]{\linewidth}
       \centering
       \includegraphics[width=\linewidth]{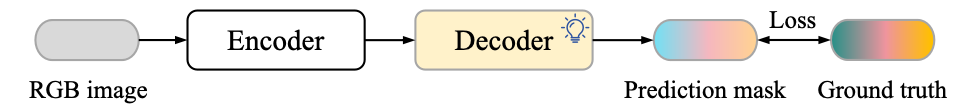}
       \\
       \footnotesize{(a) Vanilla segmentation architecture}
   \end{minipage}
   \\[1.5em] 
   \begin{minipage}[b]{\linewidth}
       \centering
       \includegraphics[width=\linewidth]{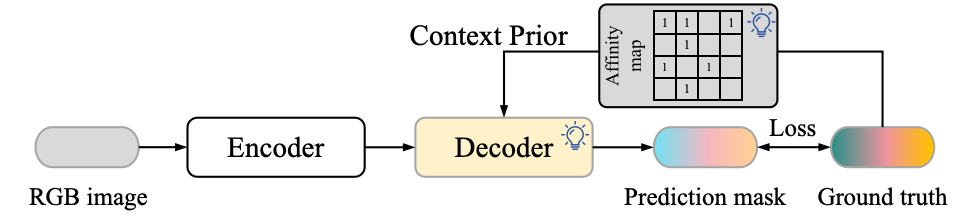}
       \\
       \footnotesize{(b) Context prior guidance from \textit{ground truth} architecture}
   \end{minipage}
   \\[1.5em]
   \begin{minipage}[b]{\linewidth}
       \centering
       \includegraphics[width=\linewidth]{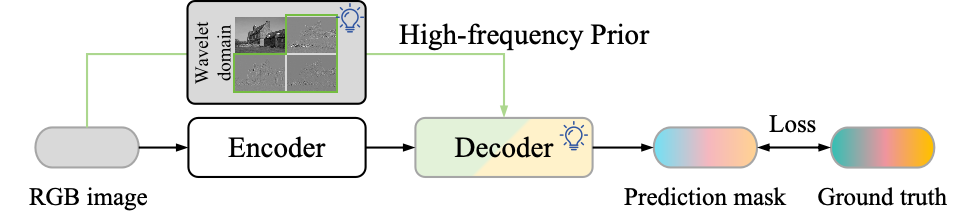}
       \\
       \footnotesize{(c) High-frequency prior guidance from \textit{input image} architecture}
   \end{minipage}
   \caption{Comparisons of the proposed high-frequency prior guidance strategy with ground-truth-based guidance and conventional vanilla segmentation architectures.}
   \label{fig:arch_comparison}
   \vspace{-2mm}
\end{figure}

To address the limitations of conventional decoder designs, an increasing number of studies have turned to Transformer-based decoders~\cite{shim2023feedformer,xu2024macformer,yeom2025u}. These methods adopt multi-head self-attention (MHSA) to extract token representations from hierarchical feature maps and iteratively refine them within the Transformer layers. Through this process, the models capture long-range dependencies and progressively broaden their receptive fields. However, a major bottleneck arises from the quadratic computational complexity of MHSA with respect to input resolution, which significantly restricts its practical deployment in dense prediction tasks. 
To alleviate this, more efficient variants of MHSA, such as local/window-based attention~\cite{shi2022ssformer,yan2024multi}, axial attention~\cite{shao2025mcanet}, and lightweight attention mechanisms~\cite{pan2022edgevits,li2023rethinking,xu2024scaseg,wang2024repvit}, have been explored.

\begin{figure*}[t]
	\centerline{\includegraphics[width=18cm]{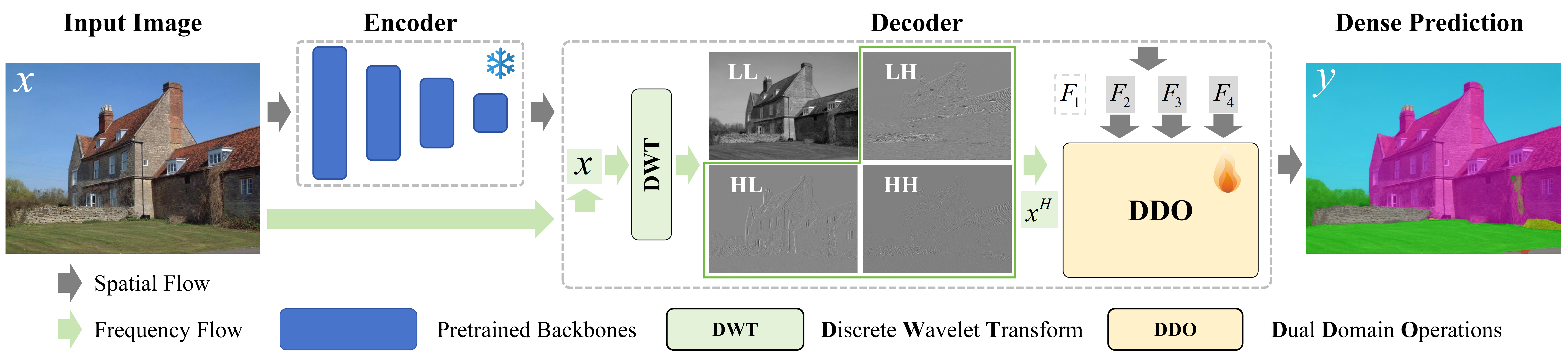}}
         \caption{Overall segmentation pipeline. The input image is first encoded into multi-level features by a pretrained backbone. These hierarchical features are then progressively enhanced and fused in our proposed dual-domain decoder.}
	\label{fig:network}
    \vspace{-2mm}
\end{figure*}
%
%
%
Although facilitating global information exchange, these attentions often overlook fine-grained local details, which are indispensable for accurate segmentation, particularly along object boundaries. From the above observations, we find that an effective and practical decoder should simultaneously possess four key capabilities: aggregating multi-scale contextual features from the encoder, modeling long-range token dependencies, preserving fine structural details, and maintaining low computational complexity to ensure efficient inference. 

In this paper, we propose an efficient decoder that effectively aggregates multi-scale features from the encoder and jointly optimizes feature refinement in both spatial and wavelet domains. 
Different from existing dual-domain approaches, our method learns high-frequency components from input images as explicit priors to guide the decoder features at different hierarchical levels.
This high-frequency structural prior compensates for detailed information often lost during the encoding process, which the decoder alone cannot fully recover, thereby improving fine-grained perception and boundary reconstruction capabilities. 
Compared to edge-based methods like~\cite{liu2024bpkd} that rely on fixed and hand-crafted Canny edges, our approach introduces a learnable mechanism that can be seamlessly integrated into the network. In addition, while some other methods~\cite{yu2020context} rely on ground-truth-generated context priors, which introduce a strong dependency on annotations, our approach eliminates the need for labels by learning structurally rich prior information directly from the image.
Fig.~\ref{fig:arch_comparison} illustrates the difference between our proposed high-frequency prior guidance and ground-truth-based guidance methods. Instead of relying on ground-truth supervision, our approach learns high-frequency components as explicit priors to reinforce boundary details.

\begin{figure*}[t]
	\centerline{\includegraphics[width=18cm]{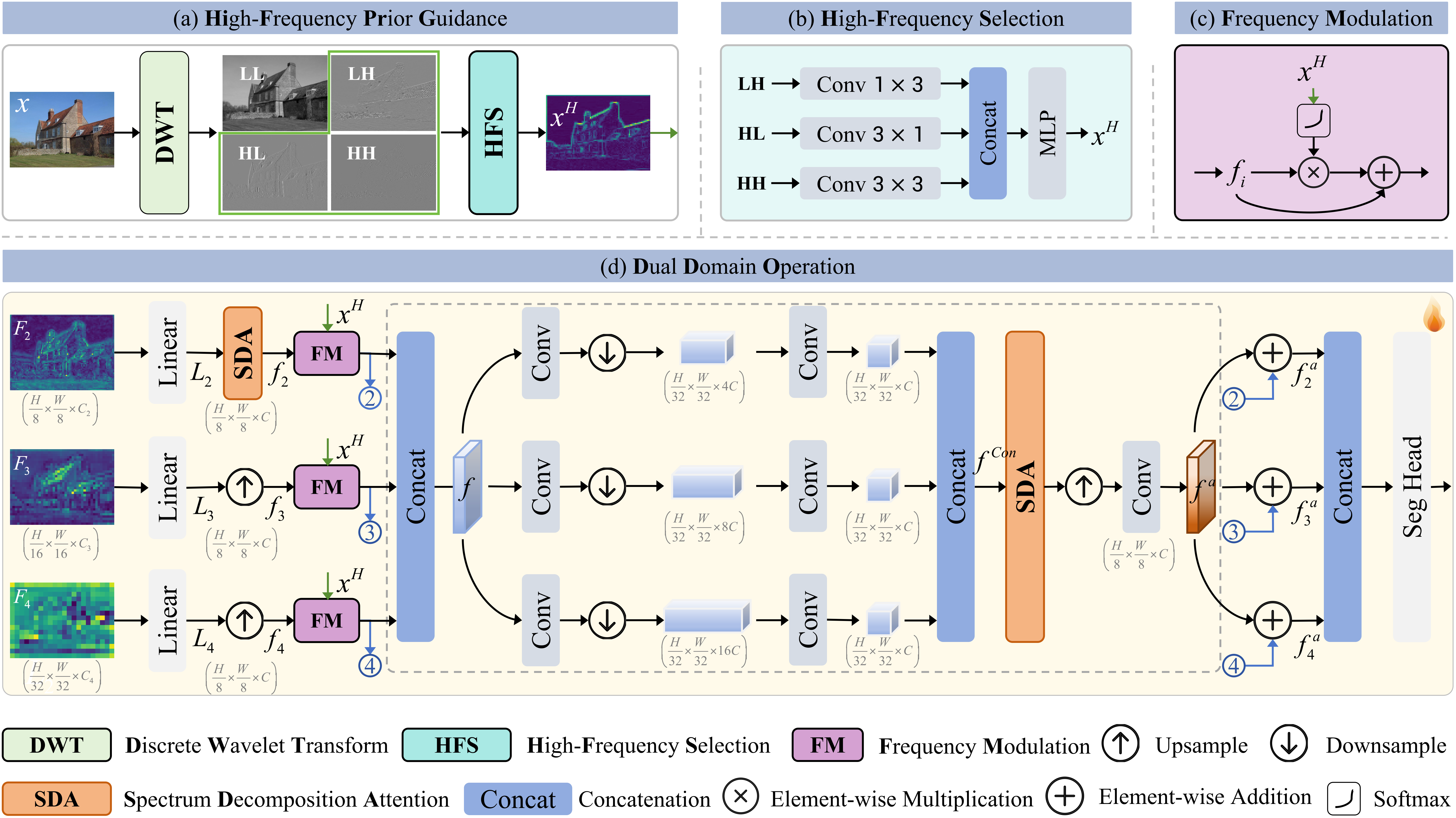}}
         \caption{Overall architecture of our proposed dual-domain decoder. Feature maps $F_2$, $F_3$, and $F_4$ are extracted from different stages of the pretrained backbone. The decoder incorporates high-frequency prior guidance and multi-scale feature enhancement to progressively refine the semantic representations and produce accurate segmentation results.}
	\label{fig:decoder}
    \vspace{-1.0em}
\end{figure*}

\begin{figure*}[t]
	\centerline{\includegraphics[width=18cm]{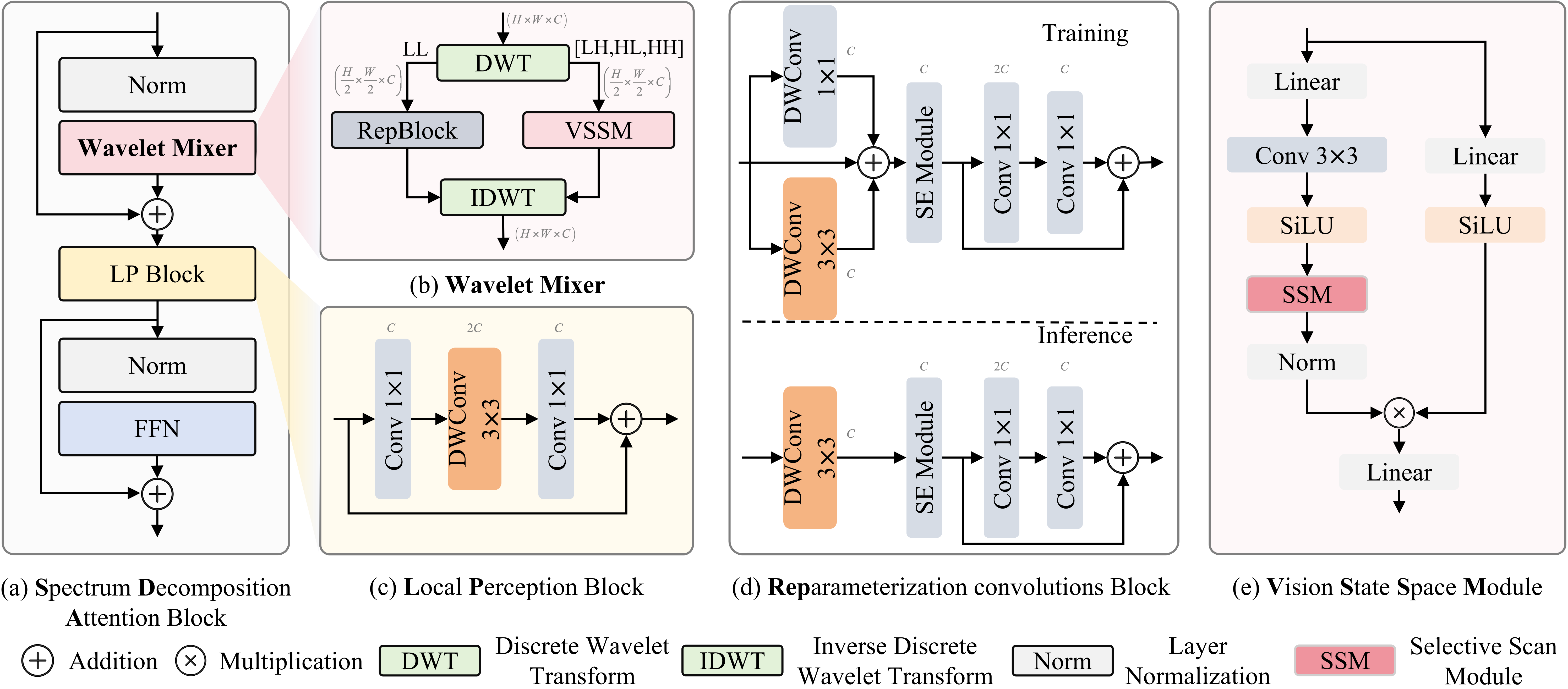}}
         \caption{The structure of our proposed Spectrum Decomposition Attention (SDA) block. The Wavelet Mixer (WM) operates in the wavelet domain to model high-frequency structural details, while the Local Perception Block (LPB) works in the spatial domain to enhance local contextual understanding. The two branches are fused to effectively combine spatial and frequency cues for improved feature representation.}
         \vspace{-1.5em}
	\label{fig:mixer}
\end{figure*}

\begin{figure}[t]
	\centerline{\includegraphics[width=9cm]{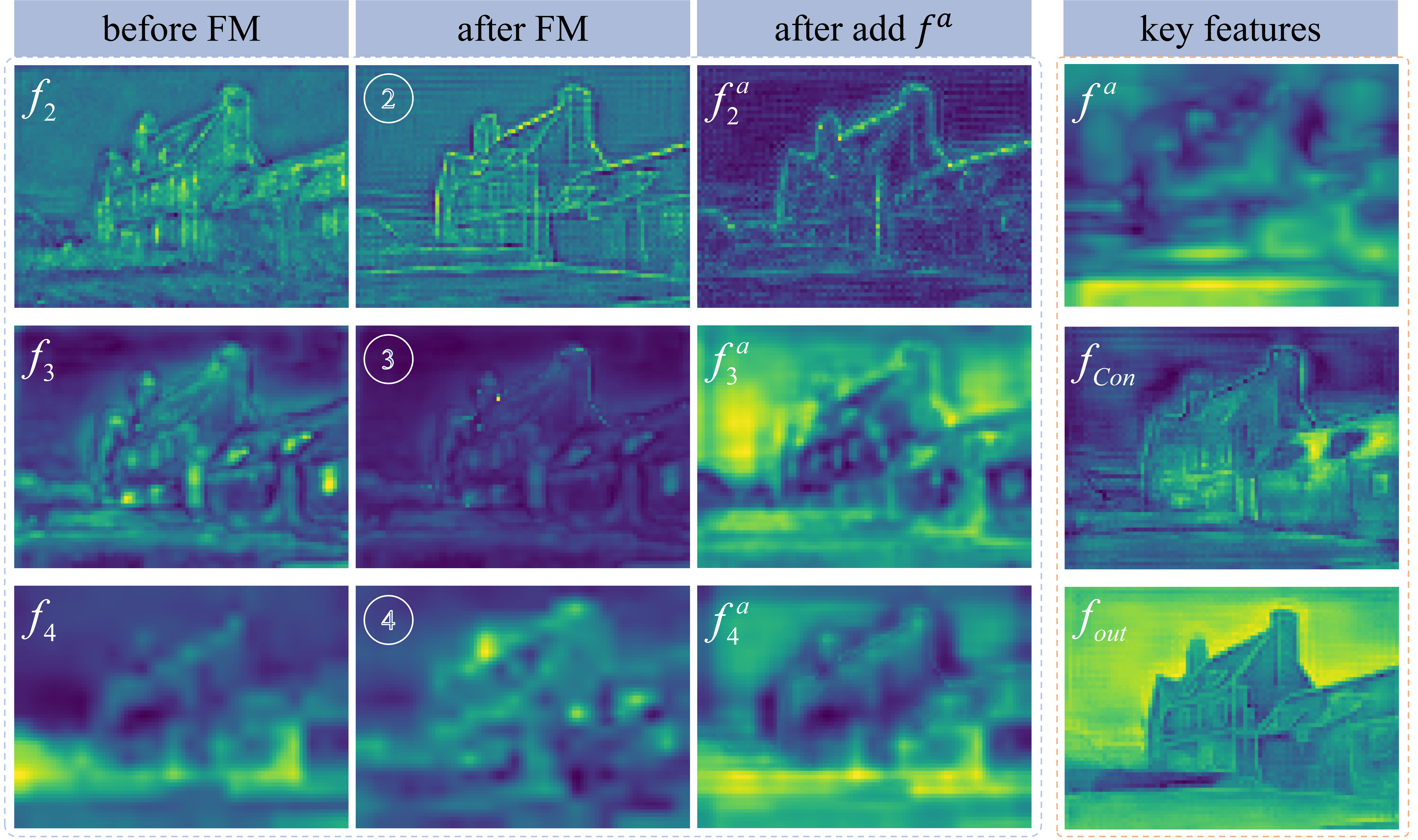}}
         \caption{Feature map visualizations before and after applying our high-frequency prior modulation (FM). The enhanced features guided by $x^H$ exhibit clearer object boundaries and richer structural details across different levels, demonstrating the effectiveness of our high-frequency prior.}
	\label{fig:feats}
\end{figure}

\begin{figure*}[t]
   \centering
   \begin{minipage}[b]{0.195\textwidth}\centering
       \includegraphics[width=\linewidth]{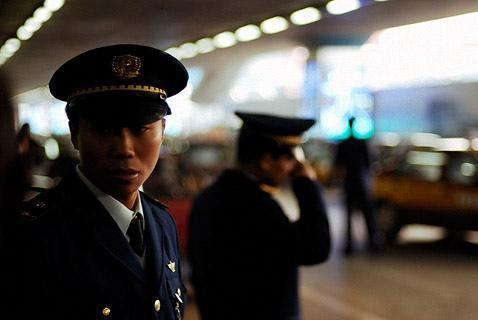}
   \end{minipage}\hfill
   \begin{minipage}[b]{0.195\textwidth}\centering
       \includegraphics[width=\linewidth]{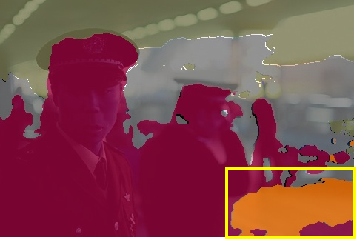}
   \end{minipage}\hfill
   \begin{minipage}[b]{0.195\textwidth}\centering
       \includegraphics[width=\linewidth]{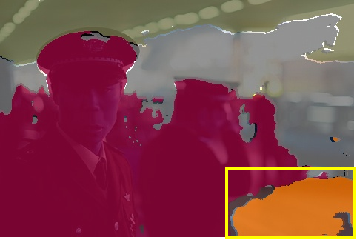}
   \end{minipage}\hfill
   \begin{minipage}[b]{0.195\textwidth}\centering
       \includegraphics[width=\linewidth]{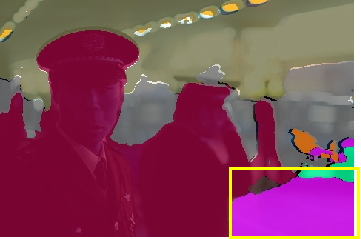}
   \end{minipage}\hfill
   \begin{minipage}[b]{0.195\textwidth}\centering
       \includegraphics[width=\linewidth]{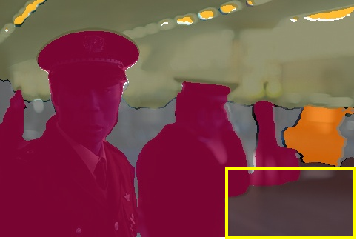}
   \end{minipage}
   \\[2pt]
   \begin{minipage}[b]{0.195\textwidth}\centering
       \includegraphics[width=\linewidth]{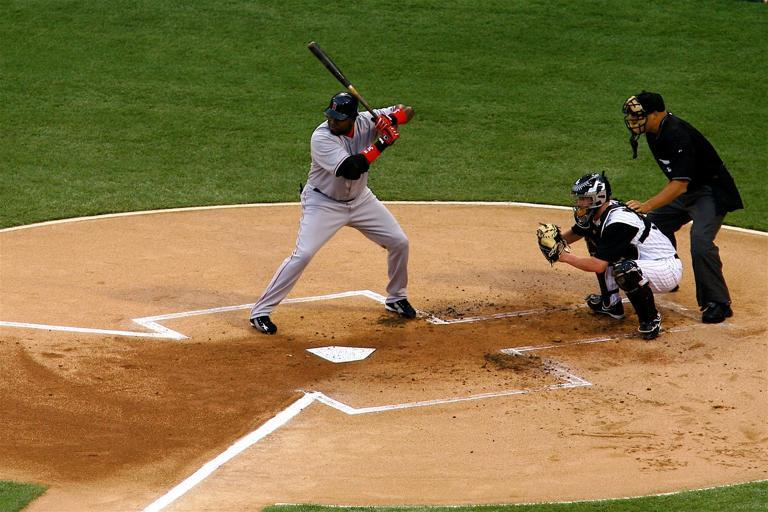}
   \end{minipage}\hfill
   \begin{minipage}[b]{0.195\textwidth}\centering
       \includegraphics[width=\linewidth]{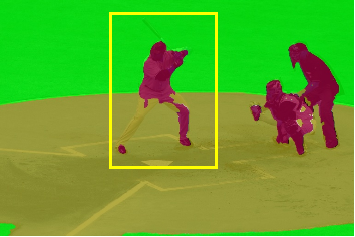}
   \end{minipage}\hfill
   \begin{minipage}[b]{0.195\textwidth}\centering
       \includegraphics[width=\linewidth]{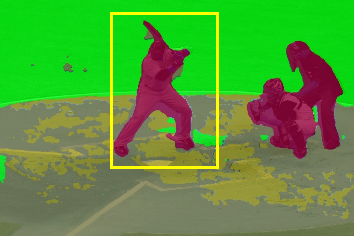}
   \end{minipage}\hfill
   \begin{minipage}[b]{0.195\textwidth}\centering
       \includegraphics[width=\linewidth]{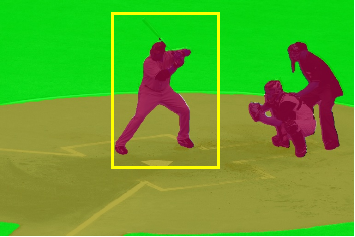}
   \end{minipage}\hfill
   \begin{minipage}[b]{0.195\textwidth}\centering
       \includegraphics[width=\linewidth]{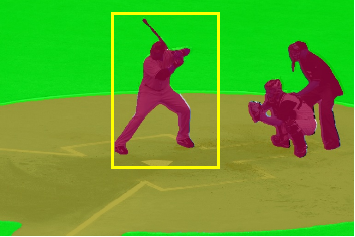}
   \end{minipage}
   \\[2pt]
   \begin{minipage}[b]{0.195\textwidth}\centering
       \includegraphics[width=\linewidth]{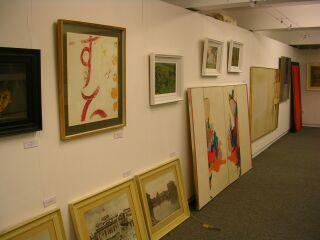}
   \end{minipage}\hfill
   \begin{minipage}[b]{0.195\textwidth}\centering
       \includegraphics[width=\linewidth]{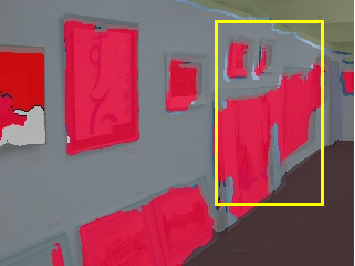}
   \end{minipage}\hfill
   \begin{minipage}[b]{0.195\textwidth}\centering
       \includegraphics[width=\linewidth]{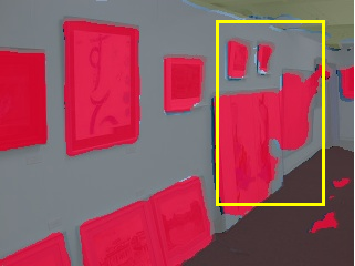}
   \end{minipage}\hfill
   \begin{minipage}[b]{0.195\textwidth}\centering
       \includegraphics[width=\linewidth]{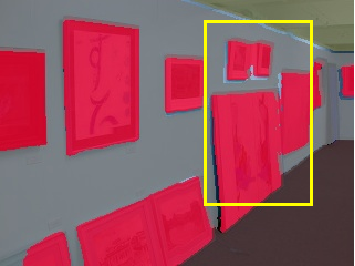}
   \end{minipage}\hfill
   \begin{minipage}[b]{0.195\textwidth}\centering
       \includegraphics[width=\linewidth]{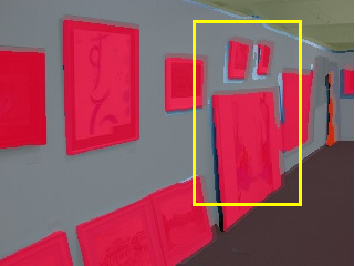}
   \end{minipage}
   \\[2pt]
   \begin{minipage}[b]{0.195\textwidth}\centering
       \includegraphics[width=\linewidth]{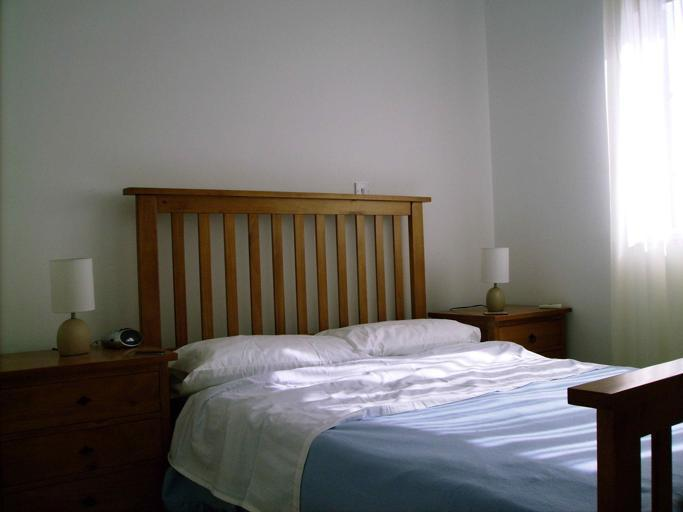}
       \vspace{0.2em}\scriptsize Input image
   \end{minipage}\hfill
   \begin{minipage}[b]{0.195\textwidth}\centering
       \includegraphics[width=\linewidth]{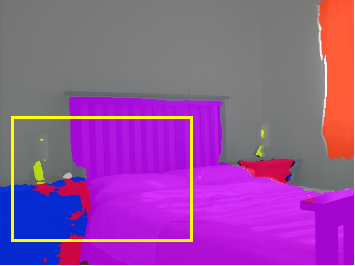}
       \vspace{0.2em}\scriptsize SegFormer~\cite{xie2021segformer}
   \end{minipage}\hfill
   \begin{minipage}[b]{0.195\textwidth}\centering
       \includegraphics[width=\linewidth]{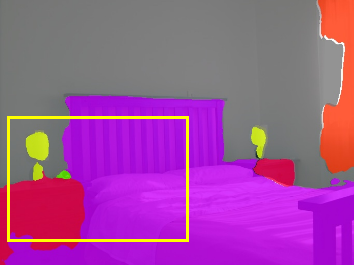}
       \vspace{0.2em}\scriptsize FeedFormer~\cite{shim2023feedformer}
   \end{minipage}\hfill
   \begin{minipage}[b]{0.195\textwidth}\centering
       \includegraphics[width=\linewidth]{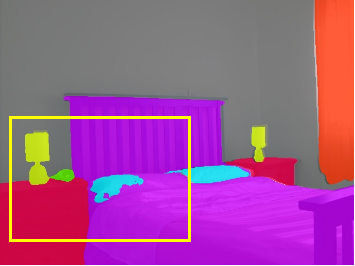}
       \vspace{0.2em}\scriptsize VWFormer~\cite{yan2024multi}
   \end{minipage}\hfill 
   \begin{minipage}[b]{0.195\textwidth}\centering
       \includegraphics[width=\linewidth]{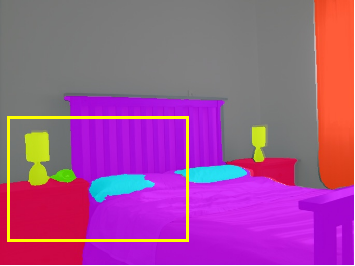}
       \vspace{0.2em}\scriptsize \textbf{WaveSeg (ours)}
   \end{minipage}
   \caption{Qualitative comparisons on ADE20K~\cite{zhou2017scene}. Columns are \textbf{Input}, \textbf{SegFormer}~\cite{xie2021segformer}, \textbf{FeedFormer}~\cite{shim2023feedformer}, \textbf{VWFormer}~\cite{yan2024multi}, and \textbf{WaveSeg (ours)}. Our method achieves more precise segmentation, particularly around object boundaries (e.g., baseball bat, pillows, lamps).}
   \label{fig: ade20k}
   \vspace{-2mm}
\end{figure*}

\begin{figure*}[t]
   \centering
   \begin{minipage}[b]{0.195\textwidth}\centering
       \includegraphics[width=\linewidth]{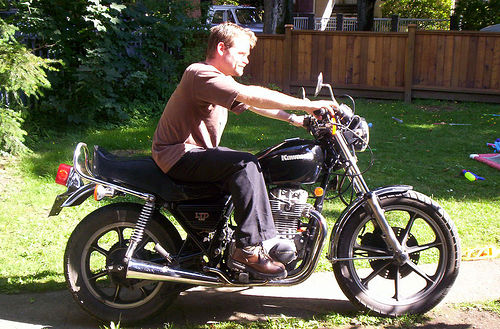}
   \end{minipage}\hfill
   \begin{minipage}[b]{0.195\textwidth}\centering
       \includegraphics[width=\linewidth]{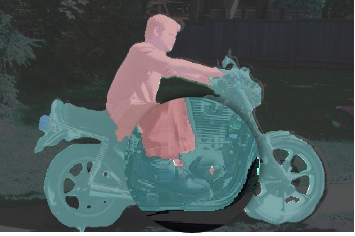}
   \end{minipage}\hfill
   \begin{minipage}[b]{0.195\textwidth}\centering
       \includegraphics[width=\linewidth]{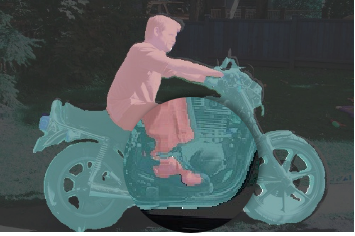}
   \end{minipage}\hfill
   \begin{minipage}[b]{0.195\textwidth}\centering
       \includegraphics[width=\linewidth]{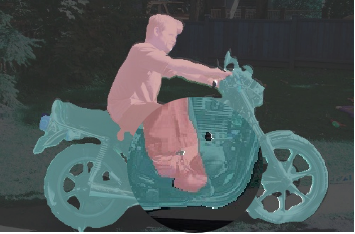}
   \end{minipage}\hfill
   \begin{minipage}[b]{0.195\textwidth}\centering
       \includegraphics[width=\linewidth]{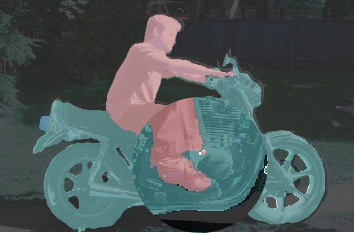}
   \end{minipage}
   \\[2pt]
   \begin{minipage}[b]{0.195\textwidth}\centering
       \includegraphics[width=\linewidth]{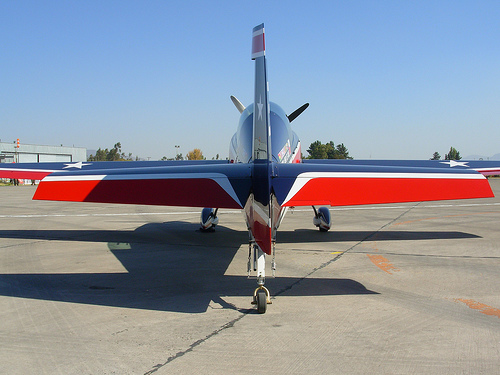}
   \end{minipage}\hfill
   \begin{minipage}[b]{0.195\textwidth}\centering
       \includegraphics[width=\linewidth]{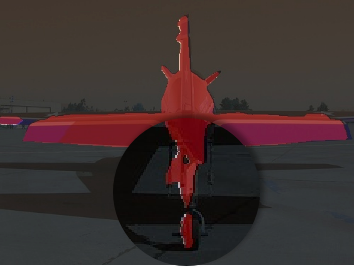}
   \end{minipage}\hfill
   \begin{minipage}[b]{0.195\textwidth}\centering
       \includegraphics[width=\linewidth]{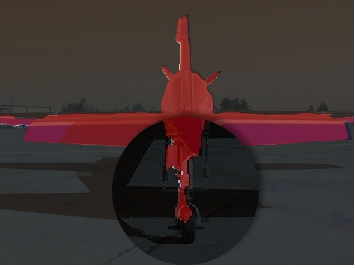}
   \end{minipage}\hfill
   \begin{minipage}[b]{0.195\textwidth}\centering
       \includegraphics[width=\linewidth]{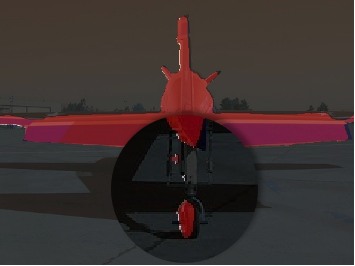}
   \end{minipage}\hfill
   \begin{minipage}[b]{0.195\textwidth}\centering
       \includegraphics[width=\linewidth]{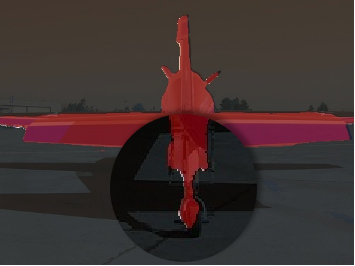}
   \end{minipage}
   \\[2pt]
   \begin{minipage}[b]{0.195\textwidth}\centering
       \includegraphics[width=\linewidth]{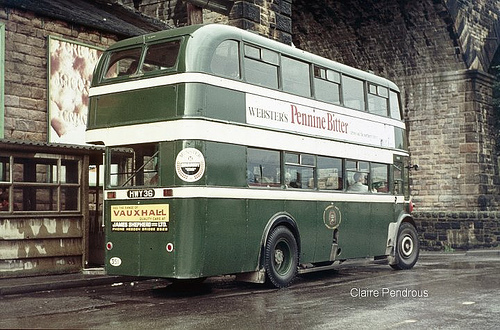}
   \end{minipage}\hfill
   \begin{minipage}[b]{0.195\textwidth}\centering
       \includegraphics[width=\linewidth]{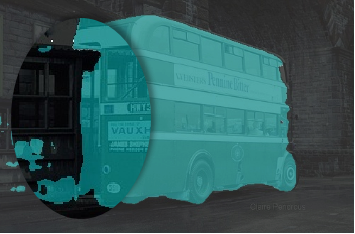}
   \end{minipage}\hfill
   \begin{minipage}[b]{0.195\textwidth}\centering
       \includegraphics[width=\linewidth]{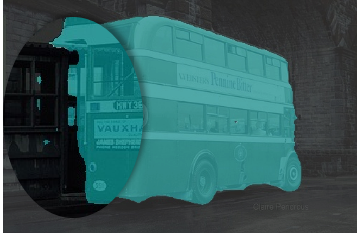}
   \end{minipage}\hfill
   \begin{minipage}[b]{0.195\textwidth}\centering
       \includegraphics[width=\linewidth]{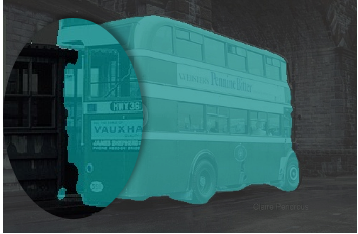}
   \end{minipage}\hfill
   \begin{minipage}[b]{0.195\textwidth}\centering
       \includegraphics[width=\linewidth]{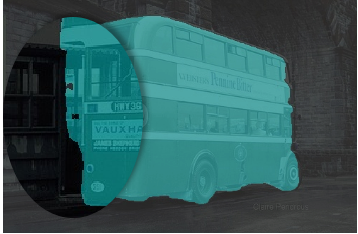}
   \end{minipage}
   \\[2pt]
   \begin{minipage}[b]{0.195\textwidth}\centering
       \includegraphics[width=\linewidth]{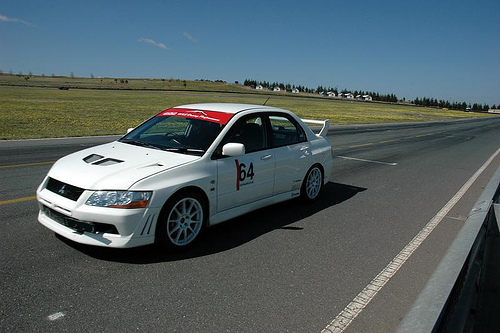}
       \vspace{0.2em}\scriptsize Input image
   \end{minipage}\hfill
   \begin{minipage}[b]{0.195\textwidth}\centering
       \includegraphics[width=\linewidth]{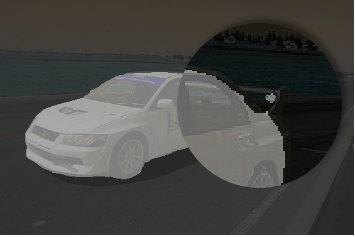}
       \vspace{0.2em}\scriptsize SegFormer~\cite{xie2021segformer}
   \end{minipage}\hfill
   \begin{minipage}[b]{0.195\textwidth}\centering
       \includegraphics[width=\linewidth]{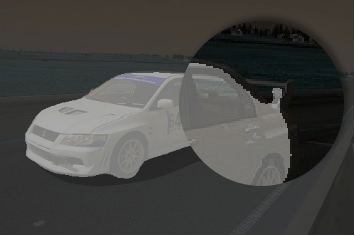}
       \vspace{0.2em}\scriptsize FeedFormer~\cite{shim2023feedformer}
   \end{minipage}\hfill
   \begin{minipage}[b]{0.195\textwidth}\centering
       \includegraphics[width=\linewidth]{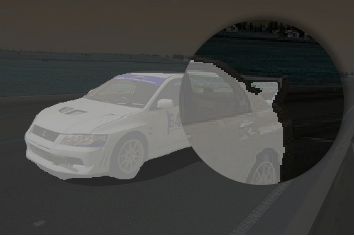}
       \vspace{0.2em}\scriptsize VWFormer~\cite{yan2024multi}
   \end{minipage}\hfill
   \begin{minipage}[b]{0.195\textwidth}\centering
       \includegraphics[width=\linewidth]{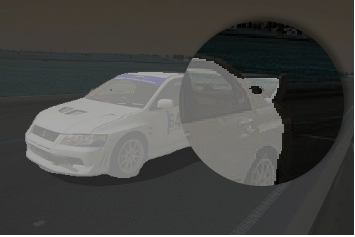}
       \vspace{0.2em}\scriptsize \textbf{WaveSeg (ours)}
   \end{minipage}
   \caption{Qualitative comparisons on PASCAL VOC2012~\cite{hoiem2009pascal}. Columns correspond to \textbf{Input}, \textbf{SegFormer}~\cite{xie2021segformer}, \textbf{FeedFormer}~\cite{shim2023feedformer}, \textbf{VWFormer}~\cite{yan2024multi}, and \textbf{WaveSeg (ours)}. The zoomed-in areas highlight regions where our method achieves more precise segmentation, especially around fine object boundaries such as feet and aircraft landing gear.}
   \label{fig: pascalvoc2012}
   \vspace{-2mm}
\end{figure*}

\begin{figure*}[t]
   \centering
   \begin{minipage}[b]{0.195\textwidth}\centering
       \includegraphics[width=\linewidth]{coco/coco1.png}
   \end{minipage}\hfill
   \begin{minipage}[b]{0.195\textwidth}\centering
       \includegraphics[width=\linewidth]{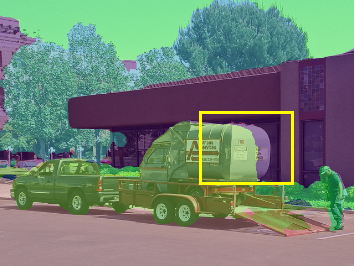}
   \end{minipage}\hfill
   \begin{minipage}[b]{0.195\textwidth}\centering
       \includegraphics[width=\linewidth]{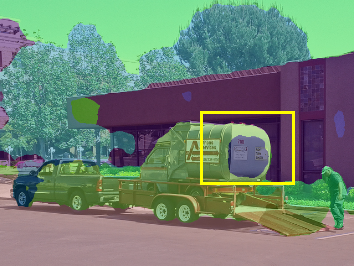}
   \end{minipage}\hfill
   \begin{minipage}[b]{0.195\textwidth}\centering
       \includegraphics[width=\linewidth]{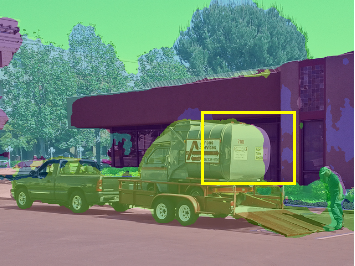}
   \end{minipage}\hfill
   \begin{minipage}[b]{0.195\textwidth}\centering
       \includegraphics[width=\linewidth]{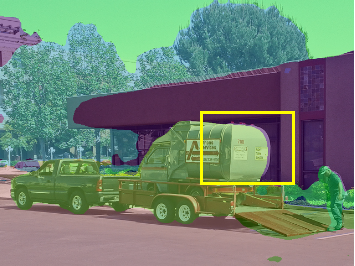}
   \end{minipage}
   \\[2pt]
   \begin{minipage}[b]{0.195\textwidth}\centering
       \includegraphics[width=\linewidth]{coco/coco2.png}
   \end{minipage}\hfill
   \begin{minipage}[b]{0.195\textwidth}\centering
       \includegraphics[width=\linewidth]{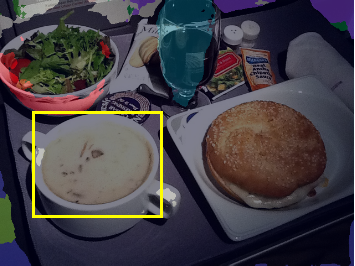}
   \end{minipage}\hfill
   \begin{minipage}[b]{0.195\textwidth}\centering
       \includegraphics[width=\linewidth]{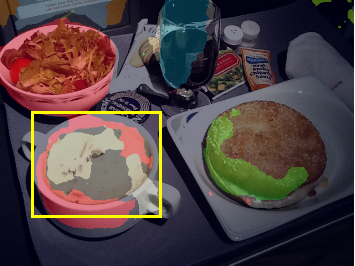}
   \end{minipage}\hfill
   \begin{minipage}[b]{0.195\textwidth}\centering
       \includegraphics[width=\linewidth]{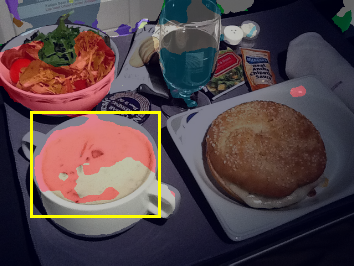}
   \end{minipage}\hfill
   \begin{minipage}[b]{0.195\textwidth}\centering
       \includegraphics[width=\linewidth]{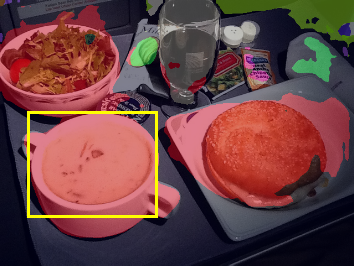}
   \end{minipage}
   \\[2pt]
   \begin{minipage}[b]{0.195\textwidth}\centering
       \includegraphics[width=\linewidth]{coco/coco3.png}
   \end{minipage}\hfill
   \begin{minipage}[b]{0.195\textwidth}\centering
       \includegraphics[width=\linewidth]{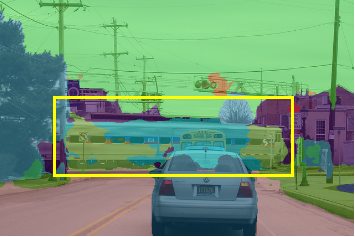}
   \end{minipage}\hfill
   \begin{minipage}[b]{0.195\textwidth}\centering
       \includegraphics[width=\linewidth]{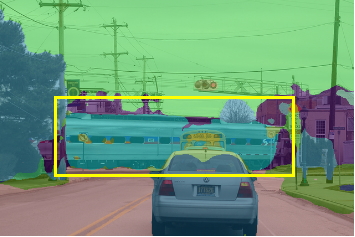}
   \end{minipage}\hfill
   \begin{minipage}[b]{0.195\textwidth}\centering
       \includegraphics[width=\linewidth]{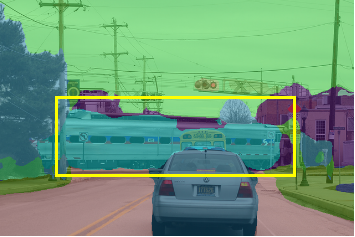}
   \end{minipage}\hfill
   \begin{minipage}[b]{0.195\textwidth}\centering
       \includegraphics[width=\linewidth]{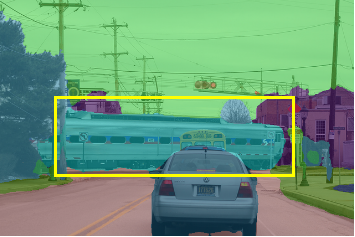}
   \end{minipage}
   \\[2pt]
   \begin{minipage}[b]{0.195\textwidth}\centering
       \includegraphics[width=\linewidth]{coco/coco4.png}
       \vspace{0.2em}\scriptsize Input image
   \end{minipage}\hfill
   \begin{minipage}[b]{0.195\textwidth}\centering
       \includegraphics[width=\linewidth]{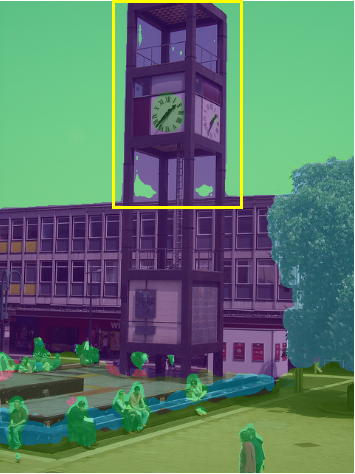}
       \vspace{0.2em}\scriptsize FeedFormer~\cite{shim2023feedformer}
   \end{minipage}\hfill
   \begin{minipage}[b]{0.195\textwidth}\centering
       \includegraphics[width=\linewidth]{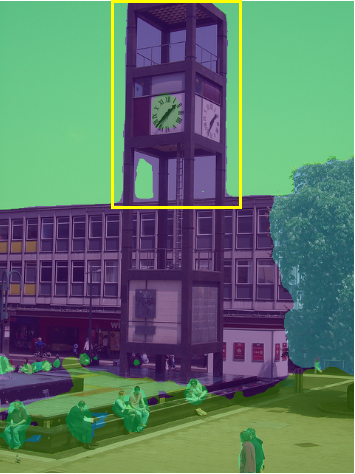}
       \vspace{0.2em}\scriptsize VWFormer~\cite{yan2024multi}
   \end{minipage}\hfill
   \begin{minipage}[b]{0.195\textwidth}\centering
       \includegraphics[width=\linewidth]{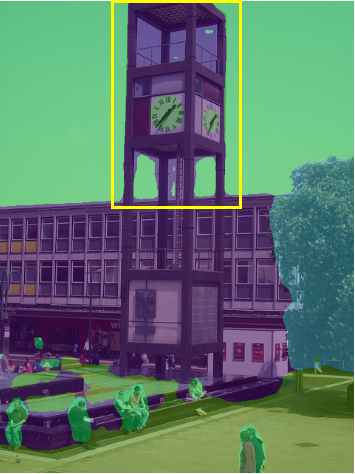}
       \vspace{0.2em}\scriptsize CCASeg~\cite{yoo2025ccaseg}
   \end{minipage}\hfill
   \begin{minipage}[b]{0.195\textwidth}\centering
       \includegraphics[width=\linewidth]{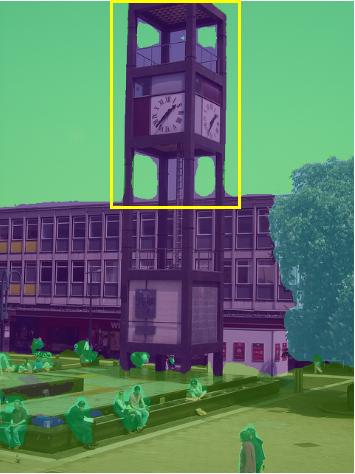}
       \vspace{0.2em}\scriptsize \textbf{WaveSeg (ours)}
   \end{minipage}
   \caption{Qualitative comparisons on COCO-Stuff 164K~\cite{caesar2018coco}. Columns correspond to \textbf{Input}, \textbf{FeedFormer}~\cite{shim2023feedformer}, \textbf{VWFormer}~\cite{yan2024multi}, \textbf{CCASeg}~\cite{yoo2025ccaseg}, and \textbf{WaveSeg (ours)}. Yellow boxes highlight regions where our method achieves more precise segmentation.}
   \label{fig: cocostuff164k}
   \vspace{-2mm}
\end{figure*}

\begin{figure*}[t]
   \centering
   \begin{minipage}[b]{0.195\textwidth}\centering
       \includegraphics[width=\linewidth]{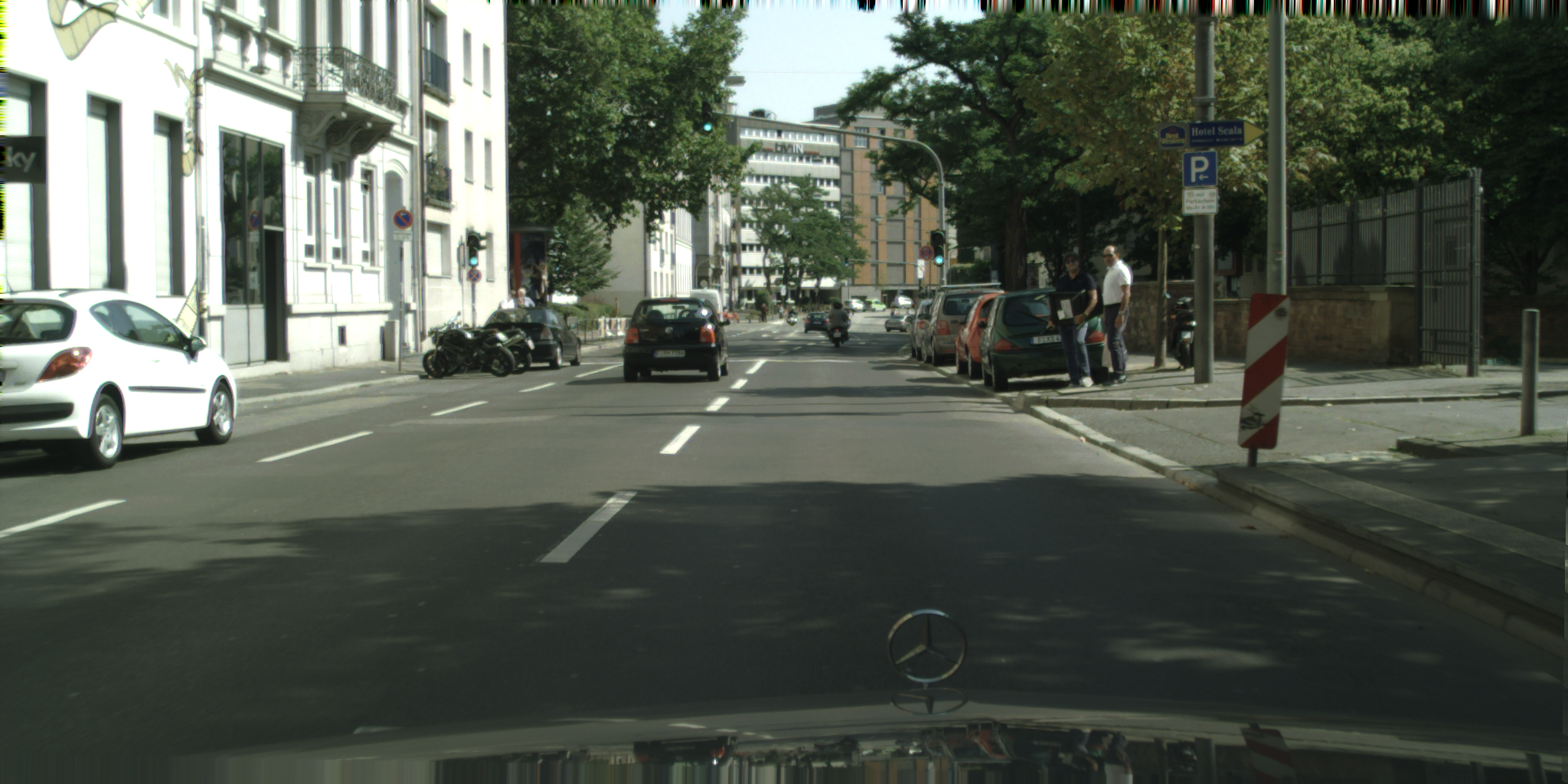}
   \end{minipage}\hfill
   \begin{minipage}[b]{0.195\textwidth}\centering
       \includegraphics[width=\linewidth]{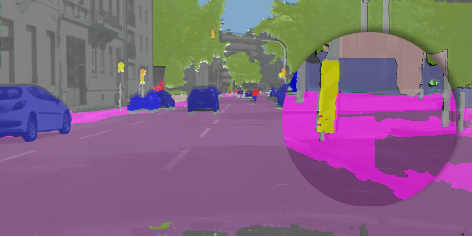}
   \end{minipage}\hfill
   \begin{minipage}[b]{0.195\textwidth}\centering
       \includegraphics[width=\linewidth]{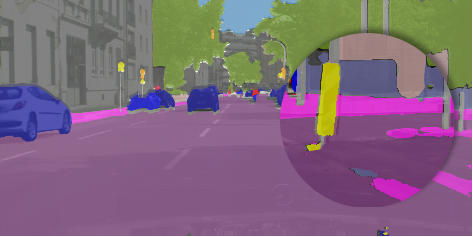}
   \end{minipage}\hfill
   \begin{minipage}[b]{0.195\textwidth}\centering
       \includegraphics[width=\linewidth]{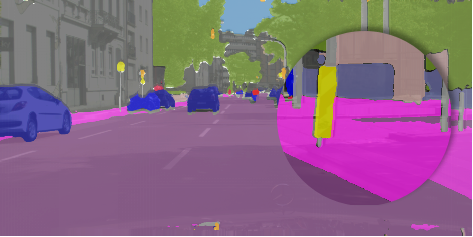}
   \end{minipage}\hfill
   \begin{minipage}[b]{0.195\textwidth}\centering
       \includegraphics[width=\linewidth]{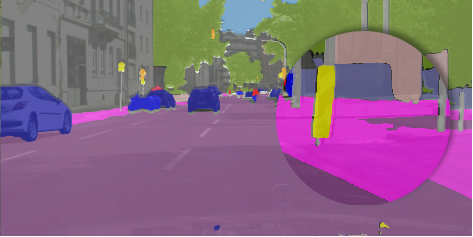}
   \end{minipage}
   \\[2pt]
   \begin{minipage}[b]{0.195\textwidth}\centering
       \includegraphics[width=\linewidth]{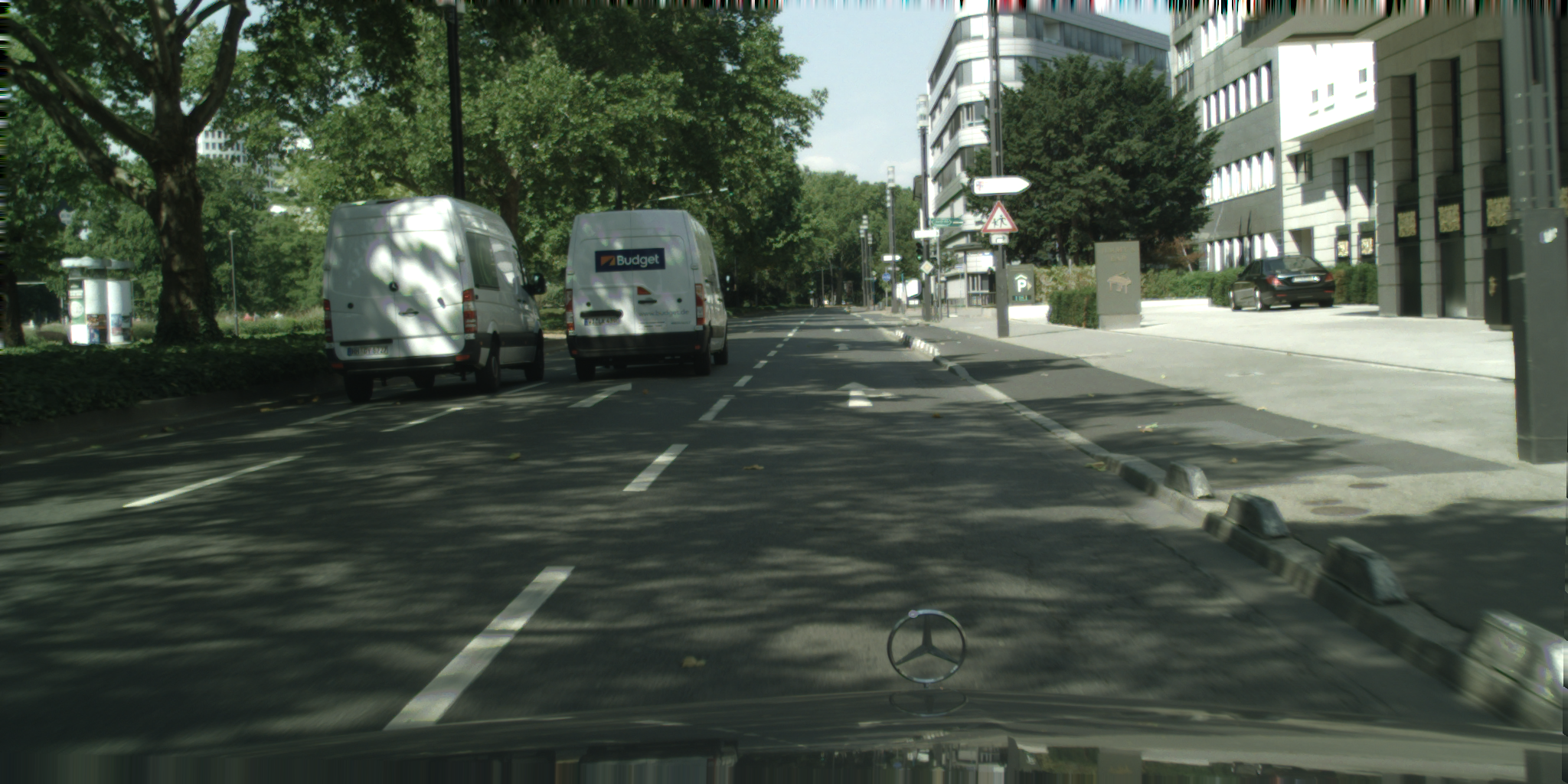}
   \end{minipage}\hfill
   \begin{minipage}[b]{0.195\textwidth}\centering
       \includegraphics[width=\linewidth]{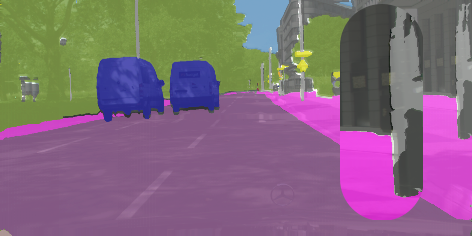}
   \end{minipage}\hfill
   \begin{minipage}[b]{0.195\textwidth}\centering
       \includegraphics[width=\linewidth]{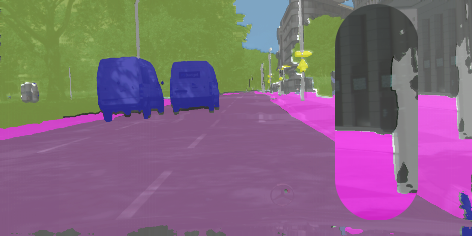}
   \end{minipage}\hfill
   \begin{minipage}[b]{0.195\textwidth}\centering
       \includegraphics[width=\linewidth]{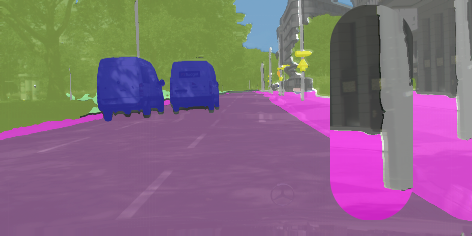}
   \end{minipage}\hfill
   \begin{minipage}[b]{0.195\textwidth}\centering
       \includegraphics[width=\linewidth]{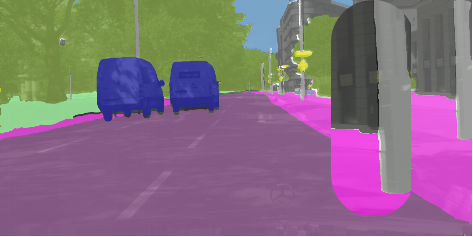}
   \end{minipage}
   \\[2pt]
   \begin{minipage}[b]{0.195\textwidth}\centering
       \includegraphics[width=\linewidth]{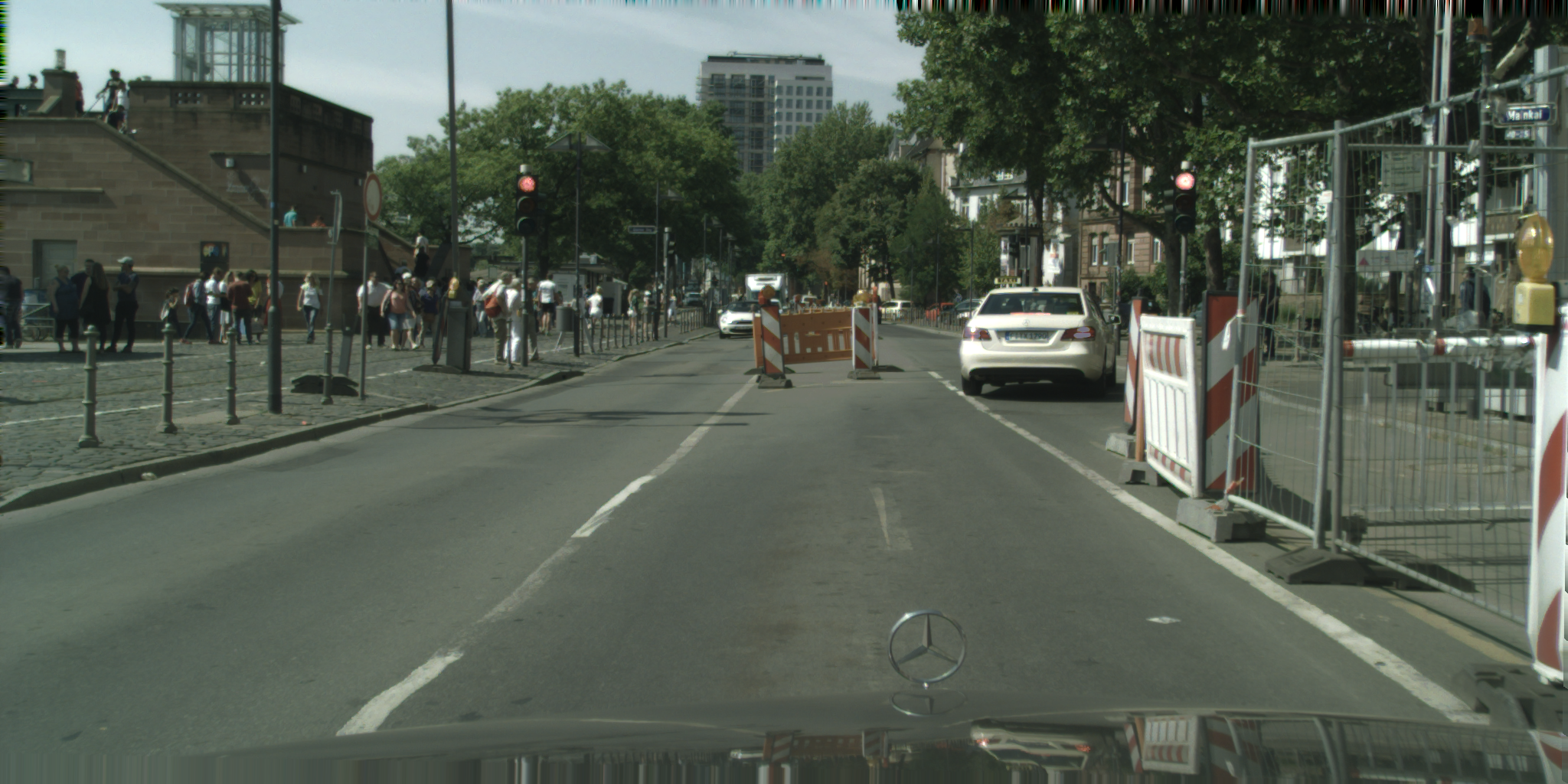}
   \end{minipage}\hfill
   \begin{minipage}[b]{0.195\textwidth}\centering
       \includegraphics[width=\linewidth]{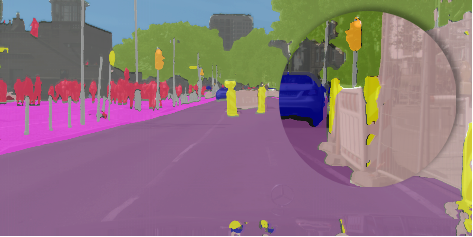}
   \end{minipage}\hfill
   \begin{minipage}[b]{0.195\textwidth}\centering
       \includegraphics[width=\linewidth]{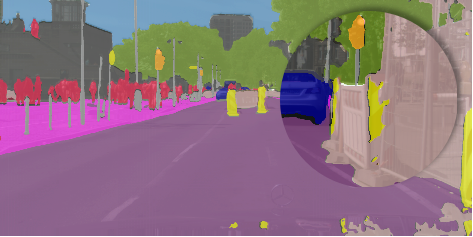}
   \end{minipage}\hfill
   \begin{minipage}[b]{0.195\textwidth}\centering
       \includegraphics[width=\linewidth]{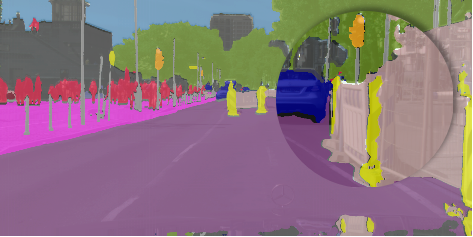}
   \end{minipage}\hfill
   \begin{minipage}[b]{0.195\textwidth}\centering
       \includegraphics[width=\linewidth]{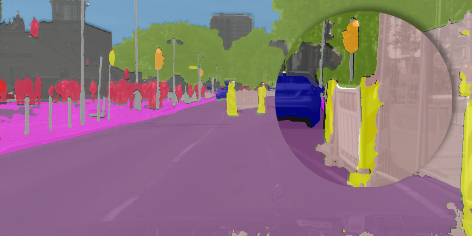}
   \end{minipage}
   \\[2pt]
   \begin{minipage}[b]{0.195\textwidth}\centering
       \includegraphics[width=\linewidth]{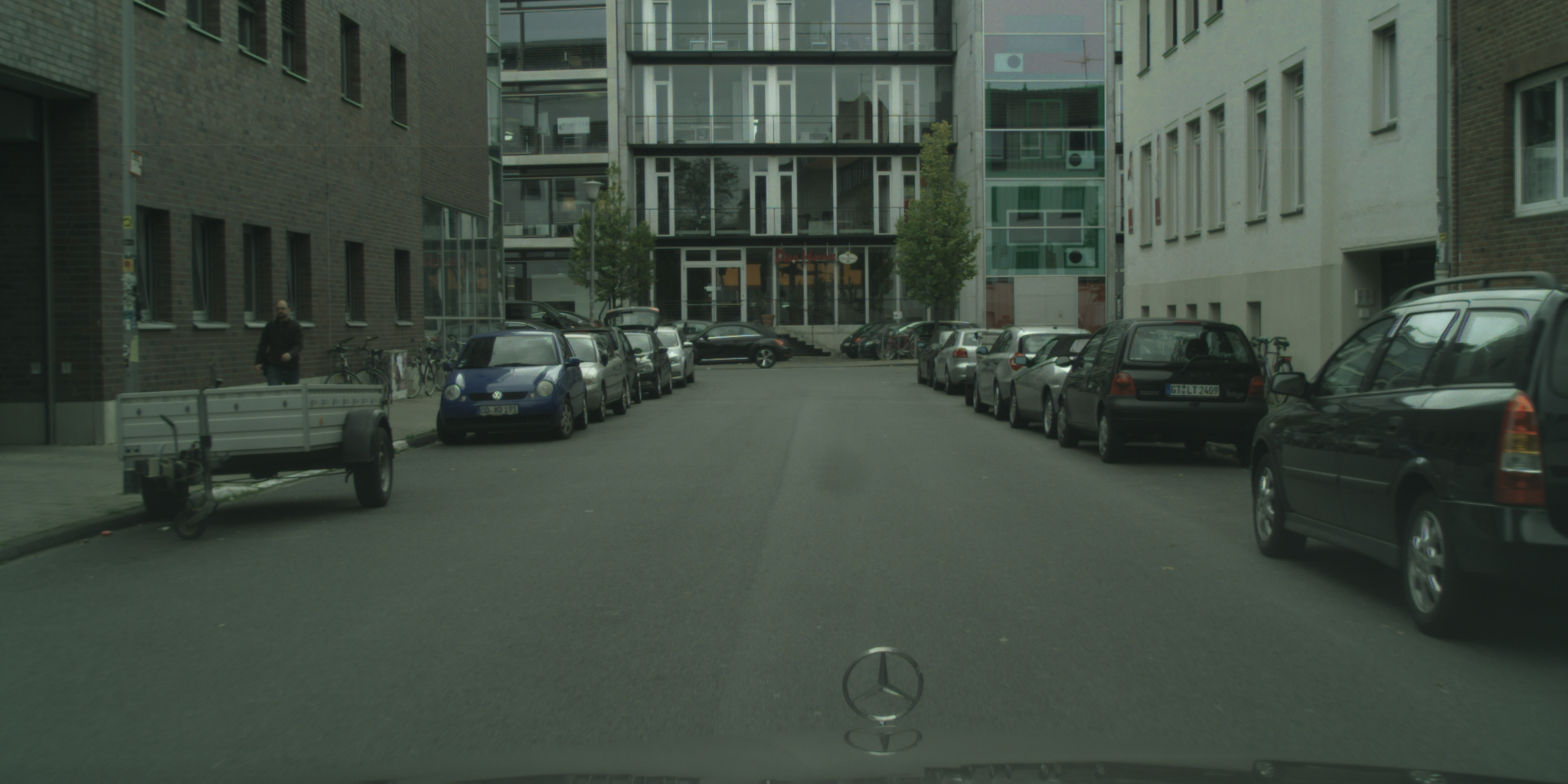}
       \vspace{0.2em}\scriptsize Input image
   \end{minipage}\hfill
   \begin{minipage}[b]{0.195\textwidth}\centering
       \includegraphics[width=\linewidth]{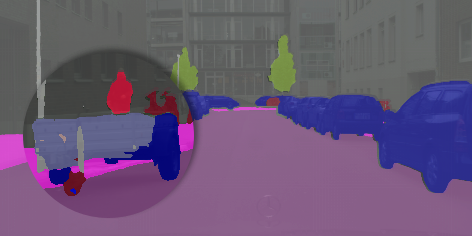}
       \vspace{0.2em}\scriptsize FeedFormer~\cite{shim2023feedformer}
   \end{minipage}\hfill
   \begin{minipage}[b]{0.195\textwidth}\centering
       \includegraphics[width=\linewidth]{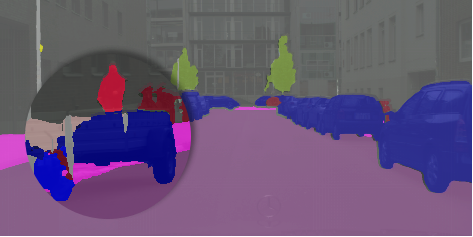}
       \vspace{0.2em}\scriptsize VWFormer~\cite{yan2024multi}
   \end{minipage}\hfill
   \begin{minipage}[b]{0.195\textwidth}\centering
       \includegraphics[width=\linewidth]{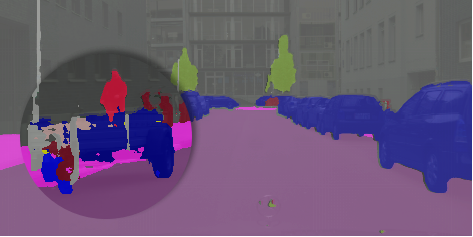}
       \vspace{0.2em}\scriptsize CCASeg~\cite{yoo2025ccaseg}
   \end{minipage}\hfill
   \begin{minipage}[b]{0.195\textwidth}\centering
       \includegraphics[width=\linewidth]{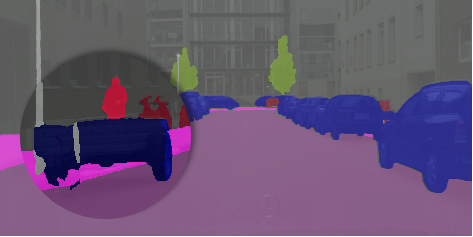}
       \vspace{0.2em}\scriptsize \textbf{WaveSeg (ours)}
   \end{minipage}
   \caption{Qualitative comparisons on Cityscapes~\cite{cordts2016cityscapes}. Columns correspond to \textbf{Input}, \textbf{FeedFormer}~\cite{shim2023feedformer}, \textbf{VWFormer}~\cite{yan2024multi}, \textbf{CCASeg}~\cite{yoo2025ccaseg}, and \textbf{WaveSeg (ours)}. The zoomed-in areas highlight regions where our method yields more precise segmentation than competing methods.}
   \label{fig: cityscapes}
   \vspace{-2mm}
\end{figure*}

Furthermore, to exploit Mamba's linear-complexity advantage, a novel Spectrum Decomposition Attention (SDA) block is introduced, where fused features are decomposed via wavelet transform and processed separately in the frequency domain. 
The high-frequency branch employs a Mamba-based module to enhance fine-grained details, while the low-frequency branch employs convolutional operations to strengthen semantic representations. Unlike MHSA and its variants, our SDA leverages Mamba to achieve truly linear global token interaction. The primary contributions can be summarized as follows:

\begin{itemize}

\item 
We propose a High-Frequency Prior Guidance (HPG) strategy, learned directly from the input, to provide fine-grained structural cues that guide the decoder features at various hierarchical levels, compensating for the detail lost during the encoder's downsampling operations.
\item 
We design a novel Spectrum Decomposition Attention (SDA) block that enhances boundary cues in the high-frequency spectrum while extracting rich contextual information from the low-frequency spectrum of multi-scale aggregated features. 
\item 
Extensive experiments on standard benchmarks demonstrate that our method consistently improves performance when integrated with various hierarchical backbones, achieving state-of-the-art results across multiple benchmark datasets.
\end{itemize}

\section{Related Works}

In this section, we briefly summarize recent representative works for image semantic segmentation and, in particular, decoders used for segmentation. 

\subsection{Semantic Segmentation}
FCN~\cite{long2015fully} pioneered in semantic segmentation by proposing to leverage CNNs in an end-to-end paradigm. Subsequently, numerous CNN-based methods have been inspired by FCN to advance further research in segmentation~\cite{yu2018bisenet,xu2023pidnet}. Then, the introduction of Vision Transformers (ViT)~\cite{dosovitskiy2020image} by SETR~\cite{zheng2021rethinking} marked a significant breakthrough, bringing this community to a new level. However, the quadratic computational complexity of MHSA posed substantial challenges, prompting a surge of interest in developing lightweight and efficient models.
For instance, Swin Transformer~\cite{liu2021swin} addressed this by restricting self-attention computation to local windows, achieving linear complexity. PVT~\cite{wang2021pyramid} further alleviated the burden by introducing spatial reduction to reduce token resolution. Meanwhile, lightweight ViTs such as MobileFormer~\cite{chen2022mobile} and MobileViT~\cite{mehta2021mobilevit} combined Transformer-based attention mechanisms with mobile-friendly architectures like MobileNet~\cite{howard2017mobilenets}. LeViT~\cite{graham2021levit} and EfficientFormer~\cite{li2022efficientformer} integrate CNNs and Transformers to better balance accuracy and inference speed. RTFormer~\cite{wang2022rtformer} proposed a two-stream architecture with cross-attention modules optimized for GPU efficiency, and SwiftFormer~\cite{shaker2023swiftformer} adopted additive attention to further reduce overhead. VWFormer~\cite{yan2024multi} employed a varying window attention scheme that enables multi-scale context modeling by performing cross-attention between local regions and windows of varying sizes. Yet, this approach suffers from limited global perception at higher resolutions, as fixed window configurations struggle to cover the full spatial extent of the feature maps. 
More recent state space models (SSMs)~\cite{zhu2024vision,liu2024vmamba,huang2024localmamba,pei2025efficientvmamba} have emerged as a compelling alternative to traditional MHSA, offering the ability to model global dependencies with linear complexity. This innovation significantly mitigates the computational burden of attention mechanisms while maintaining efficient inference. 

Despite advances in these backbones~\cite{lou2025overlock,lou2025sparx,lou2025transxnet} improving image classification, their effectiveness in segmentation remains limited, largely due to the need for detailed pixel information. To bridge this gap, we propose an efficient decoder that inherits the strong representational power of modern backbones while enhancing fine-grained prediction capabilities for semantic segmentation tasks.

\subsection{Segmentation Decoders}
Semantic segmentation critically depends on effective multi-scale representation learning, necessitating efficient decoder architectures. While established approaches like SegFormer~\cite{xie2021segformer} and UperNet~\cite{xiao2018unified} utilize simple MLPs or convolutions for feature fusion, these mechanisms fundamentally constrain comprehensive context aggregation capabilities. The paradigm shifted with DETR~\cite{zhu2020deformable}, which pioneered Transformer decoders for segmentation by replacing conventional CNN-based designs. Subsequent innovations include FeedFormer's~\cite{shim2023feedformer} cross-attention decoder for encoder feature integration, though multi-scale fusion remained suboptimal, and VWFormer's~\cite{yan2024multi} window-based attention decoder that struggles with global context modeling due to fixed window constraints. Recent frameworks like EDAFormer~\cite{zhang2024edaformer} (embedding-free Transformers with spatial reduction) and SegMAN~\cite{fu2025segman} (Mamba-enhanced decoders) demonstrate improved feature aggregation, yet persistently fail to recover fine-grained details lost during encoder downsampling.

Motivated by these limitations in existing approaches, we propose a novel strategy that directly learns high-frequency structural priors from input images to compensate for encoder-induced detail loss. This component provides explicit guidance for boundary preservation, a critical feature notably absent in prior works. Furthermore, our Spectrum Decomposition Attention (SDA) block refines aggregated multi-scale features through dual-domain processing to enhance high-frequency structural details in images. 

\section{Our Method}
\subsection{Preliminary}
\textbf{State Space Models (S4).} SSMs, which are fundamentally inspired by continuous control systems, capture the dynamic interactions between input stimulation $x(t)$
and output response $y(t)$ through linear time-invariant system frameworks. In essence, it projects the one-dimensional sequence $x(t)\in\mathbb{R}^{L}$ into $y(t)\in\mathbb{R}^{L}$ through a hidden state $h(t)\in\mathbb{R}^{N}$, Mathematically, SSMs can be formulated as linear ordinary differential equations (ODEs), detailed as follows:
\begin{equation}
    h^\prime(t) = \textbf{A}h(t) + \textbf{B}x(t),
\end{equation}
\begin{equation}
    y(t) = \textbf{C}h(t) + \textbf{D}x(t),
\end{equation}
where $\textbf{A}\in\mathbb{R}^{N\times N}$ is the evolution parameter, $\textbf{B}\in\mathbb{R}^{N\times 1}$, and $\textbf{C}\in\mathbb{R}^{1\times N}$
are the learnable projection parameters for the model state size $N$, and $\textbf{D}\in\mathbb{R}^{N}$ is the skip connection.

To enable the application of SSMs in deep learning, researchers introduced discretized variants that transform the original continuous-time formulation into discrete functions. This allows the model to operate in alignment with the sample rate of the underlying signal present in the input data $x(t)\in\mathbb{R}^{L \times D}$
, thereby facilitating integration with modern machine learning frameworks. Continuous SSMs are typically discretized using the zero-order hold method, which introduces a time scale parameter $\Delta$ 
to convert the continuous parameters 
$\textbf{A}$ and $\textbf{B}$ into their discrete forms $\bar{\textbf{A}}$ and $\bar{\textbf{B}}$. This process can be defined as follows:
\begin{equation}
    \bar{\textbf{A}} = e^{\Delta \textbf{A}}, \bar{\textbf{B}} = (\Delta \textbf{A})^{-1} (e^{\Delta \textbf{A}} - I )\cdot \Delta \textbf{B},
\end{equation}
\begin{equation}
    h^\prime(t) = \bar{\textbf{A}} h_{t-1} + \bar{\textbf{B}}x_{t},
    \label{eq_5}
\end{equation}
\begin{equation}
    y(t) = \textbf{C} h_{t} + \textbf{D}x_{t},
\end{equation}
where $\Delta \in \mathrm{R}^{D}$ and \textbf{B},\textbf{C} $\in \mathrm{R}^{D \times N}$.

\textbf{Selective State Space Models (S6).} State space models (\textit{e.g.}, S4) have been widely adopted for efficient sequence context modeling with linear computational complexity. However, their capabilities are limited by fixed parameterization, hindering flexible content-aware reasoning. To overcome these limitations, the selective state space model, S6 (\textit{i.e.}, Mamba~\cite{gu2023mamba}) was introduced. S6 enables the model to dynamically control the propagation or attenuation of information across the sequence by parameterizing the state space model as an input-dependent function of the current token's position within the sequence. Furthermore, it adopts a parallel scan approach, which echoes the efficiency benefits illustrated in Eq.~\ref{eq_5}, thereby accelerating both training and inference.

\textbf{Discrete Wavelet Transform (DWT).} DWT decomposes an input image $\mathcal{I} \in \mathbb{R}^{H \times W \times C}$ into four sub-bands with four
distinct filters representing low-frequency approximations and high-frequency details, $LL$, $LH$, $HL$, and $HH$, each of size $\frac{H}{2} \times \frac{W}{2} \times C$. This process can be expressed as follows:
\begin{equation}
    LL, LH, HL, HH = (f_{LL}, f_{LH},f_{HL},f_{HH}) \circledast \mathcal{I},
\end{equation}
where $\circledast$ denotes the convolution operator. The $LH$, $HL$, and $HH$ sub-bands retain high-frequency information — including edges and textures — across various directional components. Conversely, the $LL$ sub-band functions as a coarse approximation of the original image, capturing its low-frequency context information and supporting further recursive decomposition.

\subsection{Segmentation Encoder}
Fig.~\ref{fig:network} illustrates the overall segmentation pipeline, where input images are first encoded into multi-level features by a pretrained backbone.
These hierarchical features are then progressively enhanced and fused in our proposed decoder. 
In this work, for the encoder, we employ hierarchical encoder structures (MSCAN~\cite{guo2022segnext}, MiT~\cite{xie2021segformer}, etc.) to extract multi-scale features. 
Specifically, given an input image $I \in \mathbb{R}^{H\times W \times C}$, the pretrained backbone generates four-stage feature representations. The output of the $i$-th stage is denoted as
$F_i \in \mathbb{R}^{\frac{H}{2^i} \times \frac{W}{2^i} \times C_i}$, where $i\in{1,2,3,4}$ and $C_i$ represents the number of channels at each corresponding stage.

\subsection{Segmentation Decoder}

As shown in Fig.~\ref{fig:decoder}, to effectively capture multi-scale contextual information from the encoder outputs, we aggregate feature maps ranging from low-level $F_2$ to high-level $F_4$. 
After passing through a linear layer, 
the two semantic-rich high-level feature maps ${F_3}$ and ${F_4}$ are upsampled to a uniform spatial resolution of $\frac{H}{8} \times \frac{W}{8} \times C_i$. 
For the structure-rich intermediate-level feature $F_2$, it is further enhanced via the newly proposed Spectrum Decomposition Attention (SDA) block, which strengthens its semantic representation while preserving fine-grained boundary details. 

The above operations can be formulated as:
\begin{equation}
    {f_2} = {\mathop{\rm SDA}\nolimits} \left( {{L_2}} \right),{L_2} = \mathop{\rm Linear}\left( {{F_2}} \right),
\end{equation}
\begin{equation}
    {f_{3,4}} = \mathop{\rm Up}\left( {{L_{3,4}}} \right),{L_{3,4}} = \mathop{\rm Linear}\left( {{F_{3,4}}} \right),
\end{equation}
where $\mathop{\rm Up}$ means the upsampling operation.

Critically, a high-frequency knowledge prior extracted directly from the input image guides each feature level, explicitly compensating for the details lost during encoding. 
These refined features are then concatenated and processed with convolutional layers, and downsampled via a pixel-unshuffling operation to reduce computational cost while preserving rich semantic content. 

\textbf{High-Frequency Prior Guidance (HPG).} For the frequency flow, as depicted in Fig.~\ref{fig:decoder} (a), the input image $I$ is first resized to $\frac{1}{8}$ of its original resolution to align with the size of encoder features, and then decomposed via 
discrete wavelet transform (DWT) into four sub-bands: 
\begin{equation}
    x^{LL},x^{LH},x^{HL},x^{HH} = \mathop{\rm DWT}(x),x = {\mathop{\rm ReSize}\nolimits} \left( \mathcal{I} \right).
\end{equation}
The high-frequency sub-bands ($x^{LH}$, $x^{HL}$, and $x^{HH}$) retain images' high-frequency information, including edges and textures across various directional components, while the $x^{LL}$ sub-band functions as a coarse approximation of the original image. To enhance directional information, we design a High Frequency Selection module that applies  $1 \times 3$, $3 \times 1$, and $3 \times 3$ convolutions to $x^{LH}$, $x^{HL}$, and $x^{HH}$, respectively (see Fig.~\ref{fig:decoder} (b)), 
to suppress noise and irrelevant artifacts.

These enhanced high-frequency features are then concatenated to form a composite feature $x^H$, \textit{i.e.}, the image's high-frequency component: 
\begin{equation}
    {x^H} = \mathop{\rm MLP} ( \mathop{\rm Cat}[{\mathcal{F}^{1 \times 3}}\left( {x^{LH}} \right),{\mathcal{F}^{3 \times 1}}\left( {x^{HL}} \right),{\mathcal{F}^{3 \times 3}}\left( {x^{HH}} \right)]),
\end{equation} 
where 
$\mathop{\rm MLP}$ is Multi Layer Perception, $\mathop{\rm Cat}$ denotes the concatenation operation, ${\mathcal{F}}$ represents convolution operation.

This high-frequency prior then modulates the intermediate encoder features ($f_2$, $f_3$, and $f_4$) in the frequency domain (as shown in Fig.~\ref{fig:decoder} (c)), 
injecting high-frequency info into each level, compensating for the fine-grained details lost during the encoder's downsampling process, yielding enhanced features $f_i^{en}$ as: 
\begin{equation}
    f_i^{en} = {f_i} \otimes \psi \left( {{x^H}} \right) \oplus {f_i},
\end{equation}
where $\otimes$ and $\oplus$ denotes element-wise multiplication and addition, respectively, and $\psi$ is the Softmax function. As shown in Fig.~\ref{fig:feats}, the differences before and after applying high-frequency prior guidance are clearly visible. The enhanced features exhibit sharper and more defined boundaries, indicating better preservation of structural details.

Subsequently, the fused feature is enhanced by an SDA block, 
upsampled and added element-wise to the original features at each 
level for feature refinement as:
\begin{equation}
    f_i^a = {f^a} \oplus f_i^{en},{f^a} = {\mathcal{F}^{3 \times 3}}\left( {\mathop{\rm Up} (\mathop{\rm SDA}\left( {{f^{con}}} \right))} \right).
\end{equation}

This final features integrate 
multi-level semantics, long-range global context, local structural cues, and edge-aware information, 
enabling precise pixel-level prediction: 
\begin{equation}
    \mathcal{Y} = \mathop{\rm SegHead}\left( {{f_m}} \right),{f_m} = \mathop{\rm Cat}[f_2^a,f_3^a,f_4^a],
\end{equation}
where ${\rm SegHead}$ is the segmentation head. 

\textbf{Spectrum Decomposition Attention (SDA) Block.} As shown in Fig.~\ref{fig:mixer}, 
our proposed Spectrum Decomposition Attention (SDA) block introduces two key enhancements over conventional Multi-Head Self-Attention (MHSA): wavelet-mixer and local perception, formulated as:
\begin{equation}
    \begin{array}{l}
    y_m  = \mathop{\rm WM}(\mathop{\rm LN}(y)) + y,\\ [.5em]
    O  =  \mathop{\rm FFN}(\mathop{\rm LN}(\mathop{\rm LPB}(y_m))) + \mathop{\rm LPB}(y_m),
    \end{array}
\end{equation}
where $y$ and $O$ are the input/output features, $\rm WM$ represents Wavelet Mixer, $\rm LN$ means Layer Normalization, and $\rm LPB$ denotes Local Perception Block.

Specifically, the SDA block first decomposes the input feature via wavelet transform into high- and low-frequency components. 
The high-frequency components (${y^{LH}},{y^{HL}},{y^{HH}}$) are processed by a Mamba-based module VSSM for global frequency-domain modeling and boundary enhancement, 
while the low-frequency component (${y^{LL}}$) is refined using re-parameterization convolutions (RepBlock) to strengthen its semantic representations as:
\begin{equation}
    \begin{array}{c}
    {y^{LL}},{y^{LH}},{y^{HL}},{y^{HH}} = \mathop{\rm DWT}(y),
    \\ [.5em]
    y_{wm} = \mathop{\rm IDWT} \left( {\left( {{\mathop{\rm Rep}\nolimits} \left( {{y^{LL}}} \right), \mathop{\rm VSSM}({y^{LH}},{y^{HL}},{y^{HH}})} \right)} \right),
    \end{array}
\end{equation}
where $y_{wm}$ is the output feature of the Wavelet Mixer.
The enhanced features are then reconstructed via inverse wavelet transform (IDWT) and further refined by a local perception block to strengthen spatial details. 
This dual-domain design simultaneously enhances global context modeling, local detail preservation, and semantic understanding, making SDA a highly versatile and generalizable component.

\begin{table*}[!t]
\caption{Performance Comparisons on ADE20K~\cite{zhou2017scene}, Cityscapes~\cite{cordts2016cityscapes}, COCO-Stuff 164k~\cite{cordts2016cityscapes}, and PASCAL VOC2012~\cite{hoiem2009pascal} Datasets Using Tiny Size Version Encoders as Backbone. The Best Results are \textbf{Boldfaced}, and the Second-Best Results are \underline{Underlined}.}
\label{main results}
\begin{center}	
\scalebox{1.0}{
\begin{tabular}{lc|cc|cc|cc|cc}
\toprule
\multirow{2}{*}{Method}  & \multirow{2}{*}{Param. (M)$\downarrow$} & \multicolumn{2}{c|}{ADE20K} & \multicolumn{2}{c|}{Cityscapes} & \multicolumn{2}{c|}{COCO-Stuff 164K} & \multicolumn{2}{c}{PASCAL VOV2012}\\
&   & GFLOPs$\downarrow$ & mIoU (\%)$\uparrow$ & GFLOPs$\downarrow$ & mIoU (\%)$\uparrow$  & GFLOPs$\downarrow$ & mIoU (\%)$\uparrow$   & GFLOPs$\downarrow$ & mIoU (\%)$\uparrow$ \\ \midrule \midrule 

\multicolumn{10}{c}{MiT-B0}\\ \midrule

SegFormer~\cite{xie2021segformer}   & \underline{3.8} & 8.4  & 37.4   & 125.5  & 76.2 & 8.4 & 35.6 & 8.4 & 66.5\\
FeedFormer~\cite{shim2023feedformer}   & 4.5  & 7.8  & 39.2  & 107.4  & 77.9 & 7.8 & 39.0 & 7.8 & 68.5 \\
U-MixFormer~\cite{yeom2025u}  & 6.1  & 6.1  & 41.2  & 101.7  & \underline{79.0}  & 6.1 & \underline{40.2} & 6.1 & 70.9\\
VWFormer~\cite{yan2024multi} & \textbf{3.7} & \underline{5.8} & 38.9 & 112.4 & 77.2 & \underline{5.8} & 36.3 & \underline{5.8} & 70.6\\
MetaSeg~\cite{kang2024metaseg} & 4.1 & \textbf{3.9} & 37.9 & \underline{90.9} & 76.7 & \textbf{3.9} & 37.7 & \textbf{3.9} & 68.7\\
CCASeg~\cite{yoo2025ccaseg}  & 6.2 & 7.2 & \underline{42.6} & 115.8 & 78.7 & 7.2 & 38.8 & 7.2 & \underline{71.3}\\
\midrule
\rowcolor{blue!10}\textbf{WaveSeg (ours)} & 7.9 & 6.4 & \textbf{42.8} & \textbf{50.8} & \textbf{79.6} & 6.4 & \textbf{40.9} & 6.4 & \textbf{72.8}\\ 
\midrule 

\multicolumn{10}{c}{SegMAN-T}\\ \midrule
FeedFormer~\cite{shim2023feedformer}  & \underline{3.1}  & 7.9  & 41.3   & \textbf{35.8}  & 76.7 & 7.9 & 41.5 & 7.9 & 75.7 \\ 
U-MixFormer~\cite{yeom2025u}  & 4.4   & 7.0   & 42.9 & 47.6  & 79.4 & 7.0 & \underline{42.0} & 7.0 &  \underline{75.8}\\ 
SegMAN~\cite{fu2025segman} & 6.4 & 6.2 & 43.0 & 52.5 & 80.3 & 6.4 & 41.3 & 6.4 & 75.1\\ 
VWFormer~\cite{yan2024multi} & \textbf{2.8} & \underline{6.2} & 41.4 & 63.9 &  78.1 & \underline{6.2} & 41.4 & \underline{6.2} & 75.3\\
MetaSeg~\cite{kang2024metaseg} & 2.9 & \textbf{4.5} & 41.7 & \underline{40.2} & 77.2 & \textbf{4.5} & 41.6 & \textbf{4.5} & 75.7 \\
CCASeg~\cite{yoo2025ccaseg} & 4.2 & 6.0& 42.8 & 50.9 & 78.8 & 6.0 & 41.2& 6.0 & 75.5\\ \midrule
\rowcolor{blue!10}\textbf{WaveSeg (ours)} & 6.9 & 6.9 & \textbf{43.8} & 55.0 & \textbf{81.2} & 6.9 & \textbf{42.5} & 6.9 & \textbf{76.1}\\
\midrule 

\multicolumn{10}{c}{LVT}\\ \midrule
SegFormer~\cite{xie2021segformer}  & \textbf{3.9}  & 10.6  & 39.3   & 140.9 & 77.6  & 10.6 & 36.3 & 10.6 &72.6\\
FeedFormer~\cite{shim2023feedformer} & 4.6  & 10.0  & 41.0   & 124.6  & 78.6  & 10.0 & 37.7 & 10.0 & 73.6\\
U-MixFormer~\cite{yeom2025u}   & 6.5  & 9.1   & \underline{43.7}   & 122.1  &  \underline{79.9} & 9.1 & 38.1 & 9.1 & 75.8   \\
VWFormer~\cite{yan2024multi} & 5.3 & 14.3 & 42.3 & 194.0 & 78.9  & 14.3 & 37.3 & 14.3 & 74.5\\ 
MetaSeg~\cite{kang2024metaseg} & \underline{4.2} & \textbf{6.0} & 40.8 & \underline{106.0} & 78.1 & \textbf{6.0} & 36.8 & \textbf{6.0} & 73.2\\
CCASeg~\cite{yoo2025ccaseg} & 6.5 & 9.8 & 43.6 & 134.0 & 79.3 & 9.8 & \underline{38.4} & 9.8 &  \underline{76.3}\\
\midrule
\rowcolor{blue!10}\textbf{WaveSeg (ours)} & 8.0 & \underline{8.4} & \textbf{44.1} & \textbf{67.3} & \textbf{80.5} &  \underline{8.4} & \textbf{38.9} &  \underline{8.4} & \textbf{77.1}\\ 
\midrule 

\multicolumn{10}{c}{MSCAN-T}\\ \midrule
SegNeXt~\cite{guo2022segnext} & \textbf{4.3}   &  \underline{6.6}   & 41.1  &  \underline{56.0}  & 79.8  &   \underline{6.6} & 38.7 &  \underline{6.6} & 76.3\\
FeedFormer~\cite{shim2023feedformer}  & 5.0  & 9.3  & 43.0   & 75.6  & 80.6 & 9.3 & 39.4 & 9.3 & 74.8 \\
U-MixFormer~\cite{yeom2025u}  & 6.7   & 7.6   & 44.4 & 90.0  & 81.0 & 7.6 & 40.0 & 7.6 & 77.8\\
VWFormer~\cite{yan2024multi}  & 5.8 & 13.6 & 42.5 & 131.4 & 80.3 & 13.6 & 38.9 & 13.6 & 76.5\\
MetaSeg~\cite{kang2024metaseg} &  \underline{4.7} & \textbf{5.5} & 42.4 & \textbf{47.9} & 80.1 & \textbf{5.5} & 39.7 & \textbf{5.5} & 75.0\\
CCASeg~\cite{yoo2025ccaseg} & 7.3 & 8.2 &  \underline{44.8} & 64.9 &  \underline{81.4} & 8.2 &  \underline{40.3} & 8.2 &  \underline{78.0}\\
 \midrule
\rowcolor{blue!10}\textbf{WaveSeg (ours)} & 7.9 & 7.7 & \textbf{45.2} & 61.5 & \textbf{81.7} & 7.7 & \textbf{40.6} & 7.7 & \textbf{78.6}\\  

\bottomrule
\end{tabular}
}

\end{center}
\end{table*}

\begin{table*}[!t]
\caption{Performance Comparisons on ADE20K~\cite{zhou2017scene}, Cityscapes~\cite{cordts2016cityscapes}, COCO-Stuff 164k~\cite{caesar2018coco}, and PASCAL VOC2012~\cite{hoiem2009pascal} Datasets Using Middle-size Version Encoders as Backbone. The Best Results are \textbf{Boldfaced}, and the Second-best Results are \underline{Underlined}.}
\label{main results middle size}
\begin{center}	
\scalebox{1.0}{
\begin{tabular}{lc|cc|cc|cc|cc}
\toprule
\multirow{2}{*}{Method}  & \multirow{2}{*}{Param. (M)$\downarrow$} & \multicolumn{2}{c|}{ADE20K} & \multicolumn{2}{c|}{Cityscapes} & \multicolumn{2}{c|}{COCO-Stuff 164K} & \multicolumn{2}{c}{PASCAL VOV2012}\\
&   & GFLOPs$\downarrow$ & mIoU (\%)$\uparrow$ & GFLOPs$\downarrow$ & mIoU (\%)$\uparrow$  & GFLOPs$\downarrow$ & mIoU (\%)$\uparrow$   & GFLOPs$\downarrow$ & mIoU (\%)$\uparrow$ \\ \midrule \midrule 

\multicolumn{10}{c}{MSCAN-S}\\ \midrule
SegNeXt~\cite{guo2022segnext} & \textbf{13.9}   & \underline{15.9}   & 44.3  & \textbf{124.6}  & 81.3  &  \underline{15.9} & 41.4 & \underline{15.9} & 78.6\\
FeedFormer~\cite{shim2023feedformer}  & 17.6  & 23.6  & 46.7   & 163.0  & 81.5 & 23.6 & 42.6 & 23.6 & 77.4 \\ 
U-MixFormer~\cite{yeom2025u}  & 24.3   & 20.8   & \underline{48.4} & 154.0  & 81.2 & 20.8 & 42.9 & 20.8 & 79.2\\ 
VWFormer~\cite{yan2024multi} & \underline{15.5} & 22.5 & 46.2 & 196.0 & 81.7 & 22.5 & 41.8 & 22.5 & 79.0\\
MetaSeg~\cite{kang2024metaseg} & 16.3 & \textbf{15.3} & 45.9 & \underline{126.0} & 81.3 & \textbf{15.3} & 42.1 & \textbf{15.3} & 76.7 \\
CCASeg~\cite{yoo2025ccaseg} & 23.9 & 24.2 & 47.7 & 192.8 & \underline{82.3} & 24.2 & \underline{43.8} & 24.2 & \underline{79.6}\\ \midrule
\rowcolor{blue!10}\textbf{WaveSeg (ours)} & 18.1 & 17.8 & \textbf{48.9} & 142.3 & \textbf{82.8} & 17.8 & \textbf{44.0} & 17.8 & \textbf{80.3}\\
\midrule 

\multicolumn{10}{c}{MiT-B1}\\ \midrule
SegFormer~\cite{xie2021segformer}  & \textbf{13.7}  & 15.9  & 42.2   & 243.7 & 78.5  & 15.9 & 41.0 & 15.9 & 71.1\\
FeedFormer~\cite{shim2023feedformer}  & 17.3  & 20.7  & 44.2 & 256.0 & 79.0   & 20.7 & 42.4 & 20.7 & 71.8 \\ 
U-MixFormer~\cite{yeom2025u}  & 24.0   & 17.8   & 45.2 & 246.8  & 79.9 & 17.8 & 42.7 & 17.8 & \underline{74.4}\\ 
VWFormer~\cite{yan2024multi} & \textbf{13.7} & 13.2 & 43.2 & 296.0 & 79.0 & 13.2 & 41.5 & 13.2 & 74.0\\
MetaSeg~\cite{kang2024metaseg} & \underline{16.0} & \textbf{12.4} & 43.8 & \underline{219.0} & 78.6 & \textbf{12.4} & 42.0 & \textbf{12.4} & 73.3 \\
CCASeg~\cite{yoo2025ccaseg} & 24.1 & 25.2 & \underline{46.0} & 317.0 & \underline{80.4} & 25.2 & \underline{43.0} & 25.2 & 73.6\\ \midrule
\rowcolor{blue!10}\textbf{WaveSeg (ours)} & 17.8 & \underline{12.7} & \textbf{46.7} & \textbf{101.2} & \textbf{80.7} & \underline{12.7} & \textbf{43.5} & \underline{12.7} & \textbf{74.9}\\
\midrule 

\multicolumn{10}{c}{MiT-B2}\\ \midrule
SegFormer~\cite{xie2021segformer}  & \underline{27.5}  & 62.4  & 46.5   & 717.1 & 81.0  & 62.4 & 43.4 & 62.4 & 78.7\\
FeedFormer~\cite{shim2023feedformer}  & 29.1  & 42.7  & 48.0  & 522.7  & 81.5 & 42.7 & 44.8 & 42.7 & 78.9 \\ 
U-MixFormer~\cite{yeom2025u}  & 35.8   & 40.0   & \underline{48.2} & 515.0  & \underline{81.7} &40.0 & \underline{45.5} & 40.0 & \underline{79.4}\\ 
VWFormer~\cite{yan2024multi} & \textbf{27.4} & 46.6 & 48.1 & 469.0 & \underline{81.7} & 46.6 & 45.2 & 46.6 & 79.1\\
MetaSeg~\cite{kang2024metaseg} & 27.8 & \underline{25.2} & 46.3 & \underline{420.0} & 81.2 & \underline{25.2} & 44.5 & \underline{25.2} & 77.8 \\
CCASeg~\cite{yoo2025ccaseg} & 35.5 & 36.4 & 47.5 & 509.5 & 81.6& 36.4 & 44.6& 36.4 & 78.5\\ \midrule
\rowcolor{blue!10}\textbf{WaveSeg (ours)} & 28.9 & \textbf{20.7} & \textbf{48.3} & \textbf{165.4} & \textbf{82.0} & \textbf{20.7} & \textbf{45.8} & \textbf{20.7} & \textbf{79.8}\\
\midrule 

\multicolumn{10}{c}{SegMAN-S}\\ \midrule
FeedFormer~\cite{shim2023feedformer}  & 28.0  & 32.0  & 49.4   & \textbf{192.5}  & 82.4 & 32.0 & 46.6 & 32.0 & 82.7 \\ 
U-MixFormer~\cite{yeom2025u}  & 35.2   & 29.8   & 50.6 & 220.8  & 82.8 & 29.8 & 46.9 & 29.8 & 83.2\\ 
SegMAN~\cite{fu2025segman} & 29.4 & 25.3 & \underline{51.3} & 218.4 & \underline{83.2} & 25.3 & \underline{47.5} & 25.3 & \underline{83.8}\\  
VWFormer~\cite{yan2024multi} & \textbf{24.5} & \underline{24.1} & 48.9 & 219.1 & 81.9& \underline{24.1} & 46.2 & \underline{24.1} & 81.9\\
MetaSeg~\cite{kang2024metaseg} & \underline{26.8} & \textbf{23.9} & 49.4 & \underline{209.7} & 82.3 & \textbf{23.9} & 46.0 & \textbf{23.9} & 82.6 \\
CCASeg~\cite{yoo2025ccaseg} & 34.1 & 29.9 & 49.7 & 239.2 & 82.6 & 29.9 & 47.1& 29.9 & 83.0\\ \midrule
\rowcolor{blue!10}\textbf{WaveSeg (ours)} & 30.4 & 26.3 & \textbf{51.8} & 210.4 & \textbf{83.4} & 26.3 & \textbf{48.6} & 26.3 & \textbf{84.3}\\

\bottomrule
\end{tabular}
}

\end{center}
\end{table*}

\begin{table}[!t]
\caption{Performance Comparisons on ADE20K~\cite{zhou2017scene} with Different Backbones. The Best Results are \textbf{Boldfaced}, and the Second-best Results are \underline{Underlined}.}
\label{different backbones}
\begin{center}	
\scalebox{0.9}{
\begin{tabular}{lc|ccc}
\toprule
\multirow{2}{*}{Method}  & \multirow{2}{*}{Backbone} & \multicolumn{3}{c}{ADE20K} \\
&   & P. (M)$\downarrow$ & GFLOPs$\downarrow$ & mIoU (\%)$\uparrow$ \\ \midrule \midrule 

OCRNet~\cite{yuan2020object} & HRNet-W48 & 70.5 & 164.8 & 45.6 \\
CPT~\cite{tang2025rethinking} & ResNet-101 & 48.5 & 100.0 & 46.8 \\
Mask2Former~\cite{he2022masked} & Swin-T & 47.0 & 74.0 & 47.7 \\
MaskFormer~\cite{cheng2021per} & Swin-T & 42.0 & 55.0 & 46.7 \\
CGRSeg~\cite{ni2024context} & CGRSeg-L & \textbf{14.9} & 35.7 & 48.3 \\
EDAFormer~\cite{yu2024embedding} & EDAFormer-B & 29.4 & 29.4 & \textbf{48.9} \\
PEM~\cite{cavagnero2024pem} & STDC-2 & 21.0 & 19.3 & 45.0 \\
OffSeg~\cite{zhang2025revisiting} & OffSeg-L & 26.4 & \textbf{17.1} & \underline{48.5} \\
 \midrule
\rowcolor{blue!10}\textbf{WaveSeg (ours)} & MSCAN-S & \underline{18.1} & \underline{17.8} & \textbf{48.9} \\

\bottomrule
\end{tabular}
}

\end{center}
\end{table}

\section{Experiments}

\subsection{Datasets and Evaluation Metrics}

We conducted experiments on benchmark datasets ADE20K~\cite{zhou2017scene}, Cityscapes~\cite{cordts2016cityscapes}, COCO-Stuff 164k~\cite{caesar2018coco}, and Pascal VOC2012~\cite{hoiem2009pascal}. 
We test WaveSeg on both lightweight and large hierarchical encoder structures, including MSCAN~\cite{guo2022segnext}, MiT~\cite{xie2021segformer}, and LVT.
The experiment protocols are the same as those compared in the method’s official repository. Performance was measured using mean Intersection-over-Union (mIoU), and computational complexity was analyzed via fvcore.
ADE20K~\cite{zhou2017scene} is a challenging scene parsing dataset containing 20,210 images for training, 2,000 for validation, and 3,352 for testing, with annotations covering 150 semantic categories. Cityscapes~\cite{cordts2016cityscapes} focuses on urban driving scenes, comprising 2,975 finely annotated images for training, 500 for validation, and 1,525 for testing. The annotations include 19 semantic categories commonly seen in street-view understanding. COCO-Stuff 164K~\cite{caesar2018coco} extends the COCO dataset by adding pixel-level stuff annotations. It contains 164,062 images labeled with 172 semantic categories, making it one of the most diverse and large-scale datasets for semantic segmentation. PASCAL VOC2012~\cite{hoiem2009pascal} consists of 1,464 training, 1,449 validation, and 1,456 testing images, with pixel-wise annotations for 21 categories (20 foreground classes + background). 

\subsection{Implementation Details}
All experiments were conducted using the MMSegmentation framework~\cite{mmseg2020}. In line with the training protocol of SegFormer~\cite{xie2021segformer}, we applied data augmentation strategies, such as random horizontal flipping and scaling within a range of 0.5 to 2.0. For dataset-specific preprocessing, we used random cropping of size $512 \times 512$ for ADE20K and COCO-Stuff 164K, and $1024 \times 1024$ for Cityscapes.
We trained the models for 160K iterations using the AdamW optimizer. The batch size was set to 16 for ADE20K and COCO-Stuff 164K, and 8 for Cityscapes and PASCAL VOC2012. The initial learning rate was 1e-6, with a 1,500-iteration warm-up phase. All training runs were performed on a server with $4\times$ NVIDIA H800 GPUs.

\subsection{Quantitative Comparisons with SOTA Methods}

The main experimental results are presented in Table~\ref{main results} with lightweight/tiny version backbones and Table~\ref{main results middle size} with the larger/small version backbones (and correspondingly higher FLOPs and memory). 

As shown in Table~\ref{main results}, under the MiT-B0 backbone~\cite{xie2021segformer}, our method achieves 42.8\% mIoU on the ADE20K dataset with only 6.4 GFLOPs, marking a significant 5.4\% improvement over SegFormer~\cite{xie2021segformer} while reducing the computational cost by 23.8\%. Although MetaSeg~\cite{kang2024metaseg} exhibits lower complexity (3.9 GFLOPs), its performance lags far behind, with only 37.9\% mIoU. On the Cityscapes dataset, the FLOPs advantage becomes more pronounced, thanks to Mamba's linear complexity, which provides a substantial computational benefit at higher resolutions. Our method requires only 50.8 GFLOPs, approximately half the cost of U-MixFormer~\cite{yeom2025u}, while achieving 0.9\% higher accuracy. On the COCO-Stuff 164K dataset, our model attains 40.9\% mIoU, outperforming the recent FeedFormer by 1.9\%, with 1.4 GFLOPs fewer in computation. On the PASCAL VOC2012 dataset, compared to the closest-performing CCASeg~\cite{yoo2025ccaseg}, our method achieves a 1.5\% gain in mIoU with 0.8 GFLOPs less. Under the SegMAN-T backbone, VWFormer~\cite{yan2024multi} uses notably fewer parameters but achieves only 41.4\% mIoU on ADE20K, falling well behind our 43.8\%, highlighting a clear accuracy gap despite its compactness. Similarly, while MetaSeg~\cite{kang2024metaseg} requires just 40.2 GFLOPs on Cityscapes, it sacrifices substantial accuracy, achieving 77.2\% mIoU compared to our 81.2\%, prioritizing efficiency over precision rather than striking an optimal balance. On COCO-Stuff 164K and PASCAL VOC2012, U-MixFormer~\cite{yeom2025u} is the closest competitor to our approach, but still lags by 0.5\% and 0.3\% mIoU, respectively, while incurring higher FLOPs than our method. Similar trends are observed under the other two backbones (LVT~\cite{yang2022lite} and MSCAN-T~\cite{guo2022segnext}), further validating the generality and effectiveness of our design. All reported GFLOPs are calculated under input resolutions of $512\times512$ for ADE20K, COCO-Stuff 164K, and PASCAL VOC2012, and $1024\times2048$ for Cityscapes. 

From Table~\ref{main results middle size} we can also observe that our method demonstrates a better balance between accuracy and efficiency with middle-size backbones, especially at high input resolutions. On the Cityscapes dataset, it achieves 80.7\% mIoU with only 101.2 GFLOPs using an MiT-B1 backbone, and 82.0\% mIoU with 165.4 GFLOPs using an MiT-B2 backbone. In contrast, CCASeg~\cite{yoo2025ccaseg} requires 509.5 GFLOPs to achieve 81.6\% mIoU, while VWFormer~\cite{yan2024multi} attains 81.7\% mIoU at a cost of 469.0 GFLOPs. Across identical backbones, our method consistently delivers impressive performance while requiring comparatively fewer parameters and lower computation. The advantage is especially pronounced on large-resolution inputs such as Cityscapes~\cite{cordts2016cityscapes}, where our lightweight design stands out and achieves a superior accuracy–efficiency trade-off.

Table~\ref{different backbones} compares performance across various backbones, showing that EDAFormer~\cite{zhang2024edaformer} matches our segmentation accuracy (48.9\% mIoU) but requires 11.3M more parameters and 11.6G additional FLOPs, highlighting our model’s lightweight design. Compared to CGRSeg~\cite{ni2024context}, which has a similar parameter count, our approach uses approximately half the computation (17.8G vs. 35.7G FLOPs) while achieving a 0.6\% higher mIoU. OffSeg~\cite{zhang2025revisiting} has slightly lower computation (17.1G vs. 17.8G FLOPs) but incurs 8.3M more parameters and a 0.4\% lower mIoU. Overall, these results confirm that our method delivers a superior balance of efficiency and accuracy.

\begin{table*}[!t]
\caption{Inference Speed Comparisons on COCO-Stuff 164k~\cite{caesar2018coco} and Cityscapes~\cite{cordts2016cityscapes} Datasets Based on Pretrained Encoder MiT-B1~\cite{xie2021segformer}. The Best Results are \textbf{Boldfaced}, and the Second-best Results are \underline{Underlined}.}
\label{speed}
\begin{center}
\begin{tabular}{lc|ccc|ccc}
\toprule
\multirow{2}{*}{Model} & \multirow{2}{*}{Paramters (M)$\downarrow$} & \multicolumn{3}{c|}{COCO-Stuff 164K (Res. $512 \times 512$)} & \multicolumn{3}{c}{Cityscapes (Res. $1024 \times 2048$)} \\
 &       & GFLOPs$\downarrow$   & mIoU (\%)$\uparrow$    & Inf. Time (ms)$\downarrow$   
   & GFLOPs$\downarrow$   & mIoU (\%)$\uparrow$  & Inf. Time (ms)$\downarrow$  \\
\midrule \midrule
SegFormer~\cite{xie2021segformer}& \textbf{13.7}   & 15.9      & 41.0      & \textbf{6.02}     & 243.7        & 78.5      &  \underline{101.80}    \\
FeedFormer~\cite{shim2023feedformer}& \underline{17.3}   & 20.7      & 42.4      & 10.48      & 256.0        &  79.0     & 316.27     \\
U-MixFormer~\cite{yeom2025u}   &  24.0  & 17.8  & 42.7     & 8.32        & 246.8       & 79.9        & 148.07     \\
VWFormer~\cite{yan2024multi}& \textbf{13.7}   &  13.2       &  41.5     & 8.77      &  295.9       &  79.0     & 150.36 \\
MetaSeg~\cite{kang2024metaseg}& 16.0   & \textbf{12.4}        & 42.0      & \underline{7.44}     & \underline{219.4}        &  78.6     & \textbf{98.74} \\
CCASeg~\cite{yoo2025ccaseg}& 24.1   & 25.2        & \underline{43.0}      & 16.95      & 317.0        &  \underline{80.4}     &  167.05 \\ \midrule
\rowcolor{blue!10}\textbf{WaveSeg (ours)}& 17.8    & \underline{12.7}        & \textbf{43.5}      & 8.25      & \textbf{101.2}        &  \textbf{80.7}     & 141.06 \\
\bottomrule
\end{tabular}
\end{center}
\end{table*}

\subsection{Qualitative Comparisons with SOTA Methods}
Under identical inference settings across datasets, we visualize and compare the results of our WaveSeg with those of other methods. On the ADE20K~\cite{zhou2017scene} dataset (as shown in Fig.~\ref{fig: ade20k}), our approach produces sharper boundary delineations—for instance, in the legs of the baseball player and the precise edges of the wall-mounted frame. On the PASCAL VOC2012~\cite{hoiem2009pascal} dataset (as shown in Fig.~\ref{fig: pascalvoc2012}), it demonstrates greater sensitivity to fine structures, such as the rider’s feet, while generating smoother, straighter bus contours and preserving small details like the rear spoiler of a low-resolution car. On the COCO-Stuff 164K dataset (as shown in Fig.~\ref{fig: cocostuff164k}), WaveSeg better preserves object completeness (\textit{e.g.}, the lid of a can truck and a coffee cup) and captures intricate patterns, such as the openwork design of a clock tower. On the Cityscapes dataset~\cite{cordts2016cityscapes} (as shown in Fig.~\ref{fig: cityscapes}), it provides more coherent ground segmentation and more precise delineation of slender structures like traffic-light poles. Overall, these qualitative results consistently demonstrate two core strengths of our method: accurate modeling of object-level coherence, enabled by the proposed SDA module, and precise recovery of boundary details, guided by the high-frequency prior.

\subsection{Inference Speed Comparisons}
We conducted inference speed evaluations using the mmsegmentation benchmark setup, measuring the runtime of our model on a single NVIDIA L40 GPU using input resolutions of $512 \times 512$ and $2048 \times 1024$. The results provide a fair and reproducible comparison of computational efficiency. As shown in Table~\ref{speed}, our WaveSeg not only delivers a clear performance advantage but also achieves impressive inference speed. Specifically, at a resolution of $512 \times 512$, WaveSeg records a latency of 8.25 ms, which is highly competitive. Although SegFormer~\cite{xie2021segformer} achieves the fastest inference due to its simple convolutional decoder, its accuracy is 2.5\% lower. At $1024 \times 2048$, WaveSeg’s efficiency becomes even more apparent, reducing FLOPs by over 50\% compared to other methods, thanks to the linear-complexity Mamba architecture. While MetaSeg~\cite{kang2024metaseg} exhibits the lowest latency, its accuracy drops to 78.6\%, which is 2.1\% lower than WaveSeg’s, and its FLOPs are more than double those of our WaveSeg. Overall, Table~\ref{speed} highlights WaveSeg’s superior balance of segmentation accuracy and efficiency, establishing a new benchmark for balanced performance in semantic segmentation.

\subsection{Ablation Studies}

\begin{figure}[t]
   \centering
   \begin{minipage}[b]{0.158\textwidth}\centering
       \includegraphics[width=\linewidth]{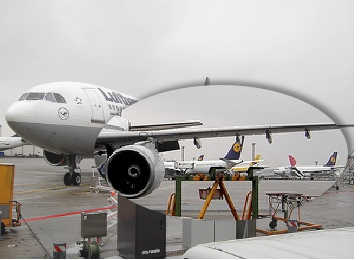}
   \end{minipage}\hfill
   \begin{minipage}[b]{0.158\textwidth}\centering
       \includegraphics[width=\linewidth]{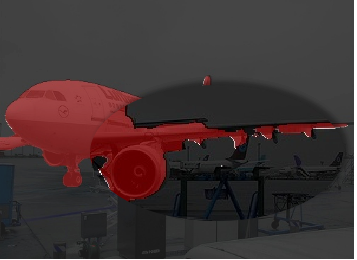}
   \end{minipage}\hfill
   \begin{minipage}[b]{0.158\textwidth}\centering
       \includegraphics[width=\linewidth]{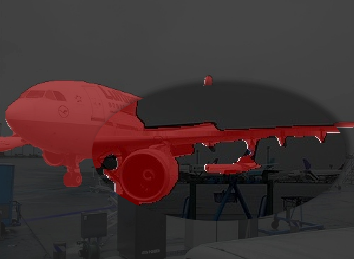}
   \end{minipage}
   \\[2pt]
   \begin{minipage}[b]{0.158\textwidth}\centering
       \includegraphics[width=\linewidth]{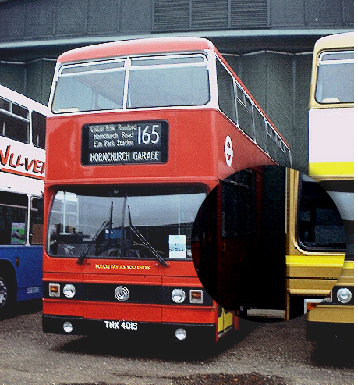}
   \end{minipage}\hfill
   \begin{minipage}[b]{0.158\textwidth}\centering
       \includegraphics[width=\linewidth]{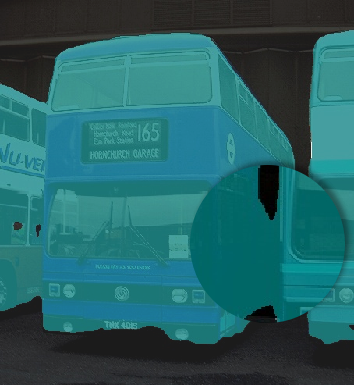}
   \end{minipage}\hfill
   \begin{minipage}[b]{0.158\textwidth}\centering
       \includegraphics[width=\linewidth]{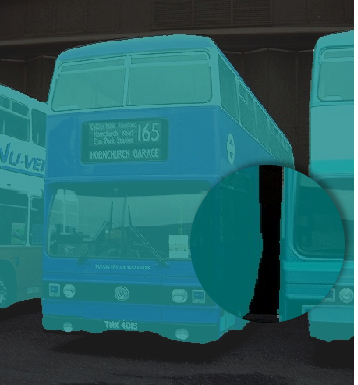}
   \end{minipage}
   \\[2pt]
   \begin{minipage}[b]{0.158\textwidth}\centering
       \includegraphics[width=\linewidth]{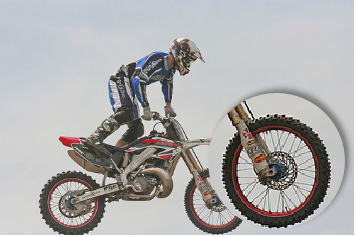}
       \vspace{0.2em}\scriptsize Input image
   \end{minipage}\hfill
   \begin{minipage}[b]{0.158\textwidth}\centering
       \includegraphics[width=\linewidth]{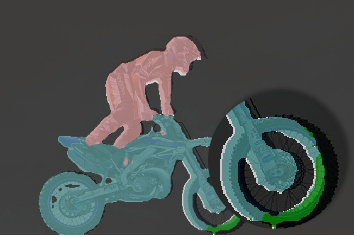}
       \vspace{0.2em}\scriptsize w/o HPG
   \end{minipage}\hfill
   \begin{minipage}[b]{0.158\textwidth}\centering
       \includegraphics[width=\linewidth]{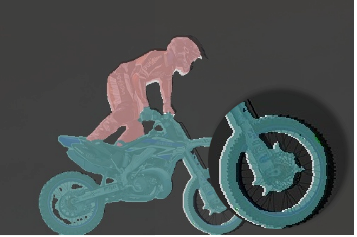}
       \vspace{0.2em}\scriptsize with HPG
   \end{minipage}
   \caption{Segmentation results obtained \textbf{with} and \textbf{without} the designed HPG strategy. The proposed HPG enhances fine-grained region segmentation, improving boundary precision.}
   \label{fig: HFPG ablation}
\end{figure}

\textbf{Effectiveness of HPG.}
To evaluate the effectiveness of our High-Frequency Prior Guidance (HPG) strategy, we conducted a series of ablation studies with and without using HPG, as summarized in Table~\ref{ablation_FPG}. 
The results demonstrate consistent performance improvements across all datasets when HPG is applied. Specifically, under a resolution of $512 \times 512$, our method achieves noticeable gains with minimal overhead, only with 0.01M additional parameters, and less than 0.5 GFLOPs. For instance, we observed a 0.6\% mIoU improvement on ADE20K, 1.1\% on COCO-Stuff 164K, and 1.2\% on PASCAL VOC2012. 
Particularly on Cityscapes, our method brings a 0.9\% gain in mIoU with an additional 3.9 GFLOPs under the shape of $1024 \times 2048$, highlighting the efficiency of HPG under high-resolution inputs. 
Furthermore, qualitative results in Fig.~\ref{fig: HFPG ablation} provide compelling evidence: the model with HPG generates significantly sharper and more accurate object boundaries, reinforcing the practical value of incorporating high-frequency priors from the input image for explicit guidance into the decoding process.

\begin{table}[!t]

\begin{center}
\caption{Ablation Studies on the Effectiveness of High-Frequency Prior Guidance (HPG) Based on the SegMAN-T~\cite{fu2025segman} Backbone.}
\label{ablation_FPG}
\scalebox{1.0}{
\begin{tabular}{c|c|cc|cc}
\toprule
\multirow{2}{*}{Model} & \multirow{2}{*}{ P. (M)$\downarrow$} & \multicolumn{2}{c|}{ADE20K} & \multicolumn{2}{c}{Cityscapes} \\  
&    & GFLOPs$\downarrow$  & mIoU$\uparrow$   & GFLOPs$\downarrow$  & mIoU$\uparrow$   \\ \midrule \midrule
w/ HPG  & 6.85   & 6.87   & \textbf{43.8}  & 54.97   & \textbf{81.2}  \\
w/o HPG & 6.84   & 6.39   & 43.2  & 51.09   & 80.3  \\ 
\midrule
\multirow{2}{*}{Model} & \multirow{2}{*}{ P. (M) $\downarrow$} & \multicolumn{2}{c}{COCO-Stuff 164K} & \multicolumn{2}{c}{PASCAL VOC2012} \\  
&    &GFLOPs $\downarrow$  & mIoU$\uparrow$   & GFLOPs$\downarrow$  & mIoU$\uparrow$   \\ \midrule
w/ HPG  & 6.85   & 6.87   & \textbf{42.5}  & 6.87   & \textbf{76.1}  \\
w/o HPG & 6.84   & 6.39   & 41.4  & 6.39   & 74.9  \\
\bottomrule
\end{tabular}
}
\end{center}
\end{table}

\begin{table}[!t]
\caption{Ablation Studies on the Effectiveness of Spectrum Decomposition Attention (SDA) Based on the SegMAN-T~\cite{fu2025segman} Backbone.}
\label{ablation_SDA}
\begin{center}
\scalebox{1.0}{
\begin{tabular}{l|cccc}
\toprule
Model Variants   & Param. (M)$\downarrow$ & GFLOPs$\downarrow$  & mIoU (\%)$\uparrow$ & FPS$\uparrow$\\
\midrule \midrule
\multicolumn{4}{c}{\textit{replace SDA}}     \\ \midrule
\multicolumn{1}{l|}{\textbf{w/ SDA}}   & 6.85   & 6.87          & \textbf{43.8}   & 46.88  \\
\multicolumn{1}{l|}{w/o SDA}  & 5.56   & 6.56   & 42.3   & 48.19     \\
\multicolumn{1}{l|}{Convolution} & 5.71  & 6.60 & 42.5   & 47.68     \\
\multicolumn{1}{l|}{CBAM}  & 5.58  & 6.56 & 42.6     & 47.47\\ \midrule
\multicolumn{4}{c}{\textit{replace WM}} \\ \midrule
\multicolumn{1}{l|}{\textbf{WM}} &  6.85  &  6.87   & \textbf{43.8}   & 46.88\\
\multicolumn{1}{l|}{MHSA} &  7.49  &  7.47  & 43.6   & 41.79  \\
\multicolumn{1}{l|}{PVT} &  8.23  &  7.38  & 43.3   & 44.35  \\
\multicolumn{1}{l|}{VSSM}&  6.86  &  6.72  & 43.2   & 47.23  \\
\bottomrule
\end{tabular}
}
\end{center}
\end{table}

\begin{table}[!t]
\caption{Ablation Studies on the Effectiveness of Integrating SDA at Different Feature Levels with the SegMAN-T~\cite{fu2025segman} Backbone.}
\label{ablation SDA2}
\begin{center}
\scalebox{1.0}{
\begin{tabular}{c|ccc|ccc}
\toprule
  & L2 & L3 & L4 & Param. (M)$\downarrow$ & GFLOPs$\downarrow$ & mIoU (\%)$\uparrow$ \\ \midrule \midrule
\multirow{10}{*}{\rotatebox{90}{SDA}}    

& \Checkmark   & \XSolidBrush   & \XSolidBrush   & 6.85  
& 6.87  & \textbf{43.8} \\ \cmidrule{2-7}
& \XSolidBrush   & \Checkmark   & \XSolidBrush   & 6.85  
& 6.09  & 43.2 \\ 
& \XSolidBrush   & \XSolidBrush   & \Checkmark   & 6.85  
& 5.89  & 42.8 \\ 
& \Checkmark   & \Checkmark   & \XSolidBrush   & 7.18  
& 7.13  & 43.5 \\
& \Checkmark   & \XSolidBrush   & \Checkmark   & 7.18  
& 6.94  & 43.3 \\
& \XSolidBrush   & \Checkmark   & \Checkmark   & 7.18  
& 6.15  & 41.4 \\
& \Checkmark   & \Checkmark   & \Checkmark   & 7.52    
& 7.20  & 42.1 \\
&  \XSolidBrush  & \XSolidBrush   & \XSolidBrush   & 6.51  
& 5.83  & 43.0  \\
\bottomrule
\end{tabular}
}
\end{center}
\end{table}

\textbf{Effectiveness of SDA.}
The effectiveness of the proposed Spectrum Decomposition Attention (SDA) block was validated through two sets of ablation studies: one evaluates the overall contribution of SDA, and the other investigates the impact of the internal Wavelet Mixer (WM). 

As shown in Table~\ref{ablation_SDA}, the performance gap between models with and without SDA is substantial. Although introducing SDA incurs a modest increase of 1.3M parameters and 0.3 GFLOPs, it yields a significant 1.5\% mIoU improvement, clearly justifying the effectiveness of the SDA module. 
Moreover, the inference speed remains largely unaffected, indicating that the proposed design achieves a favorable trade-off between accuracy and efficiency. 
This accuracy advantage is also evident when compared to conventional convolution and CBAM~\cite{woo2018cbam}, where SDA contributes an accuracy gain of 1.3\% and 1.2\%, respectively, with negligible impact on runtime.
Meanwhile, replacing WM with MHSA results in a noticeable decrease in efficiency drop, together with an increase in the GFLOPs and parameters. 
While PVT~\cite{wang2021pyramid} designs reduce some of MHSA's computational overhead, they still underperform in accuracy, lagging by 0.5\% mIoU. 
On the other hand, VSSM offers advantages in speed and reduced global modeling complexity, but it falls short in accuracy by 0.6\%, highlighting the superior balance achieved by our WM.

In addition, we explored the effectiveness of integrating SDA at different feature hierarchies generated by the encoder, as shown in Table~\ref{ablation SDA2}. Our experiments reveal that integrating SDA at $L_2$ yields the most significant performance gain. Applying SDA individually at $L_3$ or $L_4$, or simultaneously at both levels, 
even leads to degradation. Specifically, applying SDA at $L_4$ alone results in a 1.0\% drop in mIoU, while using it at both $L_3$ and $L_4$ causes a significant decline of 2.4\%, even under the same configuration.

These results further validate the design motivation of SDA: it is intended to enhance the modeling of boundary-level details through wavelet-domain representations. However, $L_3$ and $L_4$ features suffer from significant spatial detail loss due to deeper encoding stages, making them less suitable for fine-grained enhancement. Conversely, $L_1$ is too shallow and lacks sufficient semantic information. The $L_2$ level strikes a desirable balance—it retains relatively rich spatial detail while also capturing sufficient semantic context, making it the optimal point for SDA integration.

\subsection{Limitations and Future Work}
Despite achieving a superior balance between efficiency and accuracy, our method has room for improvement, particularly in the wavelet-domain Mamba block, where the fixed selective-scan order may underutilize cross-patch correlations due to Mamba’s inherently sequential nature. In the future, we aim to implement an adaptive and learnable scan order—such as policy-driven or entropy-guided traversal—that maintains Mamba’s linear complexity while aligning the sequence order with image structure. 

\section{Conclusion}

In this paper, we have proposed a novel decoder architecture named WaveSeg. Unlike previous approaches relying solely on encoder-derived features for interaction and fusion, our method introduces an explicit high-frequency prior derived directly from the input image, 
enabling a more effective recovery of fine-grained details lost during encoding. 
Compared to ground-truth-supervised prior guidance, our approach is more flexible and annotation-free. 
Furthermore, our Spectrum Decomposition Attention (SDA) block 
applies Mamba-based modeling to enhance boundary awareness in high-frequency components, 
while leveraging convolutional operations on low-frequency components to extract rich contextual semantics. 
Extensive experiments have demonstrated WaveSeg's effectiveness and compatibility with various hierarchical backbones.

\bibliographystyle{IEEEtran} 
\bibliography{ref}

\end{document}